\documentclass[11pt,a4paper]{article}
\usepackage{times,latexsym}
\usepackage{url}
\usepackage[T1]{fontenc}
\usepackage{tabularx}
\usepackage{graphicx}
\usepackage{hyperref}
\usepackage{subcaption}
\usepackage{makecell}

\newcommand{\hf}[2]{\raisebox{-2.2pt}{\includegraphics[scale=0.09]{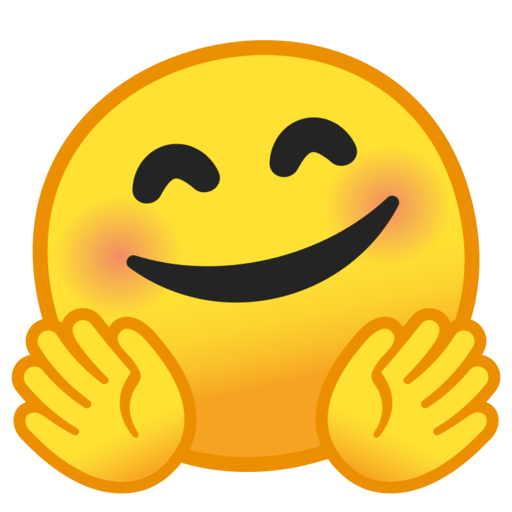}}~\href{#1}{\texttt{#2}}}
\newcommand{\opus}[2]{\raisebox{-2.2pt}{\includegraphics[scale=0.3]{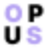}}~\href{#1}{\texttt{#2}}}

\usepackage[acceptedWithA]{tacl2021v1}
\usepackage[table]{xcolor}

\definecolor{BibleCol}{HTML}{ffd700}
\newcommand{\Bible}{\colorbox{BibleCol}{Bible}}
\newcommand{\BibleC}{\cellcolor{BibleCol}}

\definecolor{GovernmentCol}{HTML}{d4a1fc}
\newcommand{\Government}{\colorbox{GovernmentCol}{Government}}
\newcommand{\GovernmentC}{\cellcolor{GovernmentCol}}

\definecolor{HealthCol}{HTML}{ff9f96}
\newcommand{\Health}{\colorbox{HealthCol}{Health}}
\newcommand{\HealthC}{\cellcolor{HealthCol}}

\definecolor{JailbreakCol}{HTML}{808ce4}
\newcommand{\Jailbreak}{\colorbox{JailbreakCol}{\color{white}Jailbreak}}
\newcommand{\JailbreakC}{\cellcolor{JailbreakCol}\color{white}}

\definecolor{LiteratureCol}{HTML}{ffa500}
\newcommand{\Literature}{\colorbox{LiteratureCol}{Literature}}
\newcommand{\LiteratureC}{\cellcolor{LiteratureCol}}

\definecolor{MathsCol}{HTML}{d2691e}
\newcommand{\Maths}{\colorbox{MathsCol}{\color{white}Maths}}
\newcommand{\MathsC}{\cellcolor{MathsCol}\color{white}}

\definecolor{NewsCol}{HTML}{40e0d0}
\newcommand{\News}{\colorbox{NewsCol}{News}}
\newcommand{\NewsC}{\cellcolor{NewsCol}}

\definecolor{NLLBCol}{HTML}{9dcbe2}
\newcommand{\NLLB}{\colorbox{NLLBCol}{NLLB}}
\newcommand{\NLLBC}{\cellcolor{NLLBCol}}

\definecolor{OpenSubtitlesCol}{HTML}{aaee8d}
\newcommand{\OpenSubtitles}{\colorbox{OpenSubtitlesCol}{OpenSubtitles}}
\newcommand{\OpenSubtitlesC}{\cellcolor{OpenSubtitlesCol}}

\definecolor{WikipediaCol}{HTML}{909090}
\newcommand{\Wikipedia}{\colorbox{WikipediaCol}{\color{white}Wikipedia}}
\newcommand{\WikipediaC}{\cellcolor{WikipediaCol}\color{white}}

\definecolor{FloresPlusCol}{HTML}{f191d2}
\newcommand{\FloresPlus}{\colorbox{FloresPlusCol}{FLORES+}}
\newcommand{\FloresPlusC}{\cellcolor{FloresPlusCol}}

\usepackage{colortbl}
\usepackage{tacl2021v1}
\usepackage{tabularx}
\usepackage{booktabs}
\usepackage{multirow}
\usepackage{url}
\usepackage{xspace,mfirstuc,tabulary}

\newif\iftaclinstructions
\taclinstructionsfalse % AUTHORS: do NOT set this to true
\iftaclinstructions
\renewcommand{\confidential}{}
\renewcommand{\anonsubtext}{(No author info supplied here, for consistency with
TACL-submission anonymization requirements)}
\newcommand{\instr}
\fi

\iftaclpubformat % this "if" is set by the choice of options

\else

\fi

\usepackage{xcolor, soul, todonotes} 
\definecolor{kwcolor}{HTML}{1696d2}

\newcommand{\kwtodo}[1]{\todo[color=kwcolor]{$_{kw}$ {\footnotesize #1}}}
\definecolor{iwcolor}{HTML}{ca5800}
\newcommand{\isuru}[1]{\textcolor{iwcolor}{$_{isuru}$[#1]}}
\newcommand{\isurutodo}[1]{\todo[color=iwcolor]{$_{isuru}$ {\footnotesize #1}}}
\definecolor{ndscolor}{HTML}{34498b}
\newcommand{\nds}[1]{\textcolor{ndscolor}{$_{nds}$[#1]}}

\defcitealias{parliament2022constitution}{Parliament}
\defcitealias{department2024Census}{DCS SL}
\newcommand{\aliascite}[1]{(\citetalias{#1}, \citeyear{#1})}
\newcommand{\dg}{\rlap{\,$^\dagger$}}   % table comment

\newcommand{\EnSiTa}{\texttt{EnSiTa}}

\title{\EnSiTa{} - A Trilingual Multi-Domain Parallel Dataset and Benchmark for Domain-Specific Machine Translation}

\author{
   Surangika Ranathunga$^\diamond$,
   Nisansa de Silva$^\dagger$,
   Aloka Fernando$^\dagger$, 
   Kavindu Warnakulasuriya$^\ddagger$,\\
   \textbf{Isuru Wijesiri$^\star$,}
   \textbf{Menan Velayuthan$^\oplus$,}
   \textbf{Charitha Rathnayaka$^\bigtriangledown$,}
   \textbf{Thivaharan Varatharajan,}\\
  \textbf{Sajeevi Silva},
   \textbf{Piumi Kandanaarachchi$^\amalg$},
   \and
   \textbf{Uthayasanker Thayasivam$^\dagger$}  
   \\
  \ \\
   $^\diamond$Massey University \quad 
   $^\dagger$University of Moratuwa \quad 
   $^\ddagger$National University of Singapore \\ 
   $^\star$WSO2 \quad 
   $^\oplus$Utrecht University \quad 
   $^\bigtriangledown$Rowan University \quad 
   $^\amalg$District General Hospital, Hambantota
  \\
    \texttt{s.ranathunga@massey.ac.nz} \quad
    \texttt{NisansaDdS@cse.mrt.ac.lk}
 } 

\date{}

\begin{document}
\maketitle
\begin{abstract}
{Machine Translation (MT) for low-resource languages remains far behind that of high-resource languages, and the gap is widest in specialised domains, where parallel data is scarce or entirely absent. We present \EnSiTa, a trilingual multi-domain parallel dataset and benchmark for English, Sinhala and Tamil. \EnSiTa{} provides human post-edited training data for seven domains, plus manually translated test sets for those and one additional domain, all produced by professional translators under a multi-year, rigorously quality-controlled process. Using this dataset, we conduct an extensive study of domain-specific MT for all six language directions, fine-tuning a from-scratch Transformer, a pre-trained translation model (NLLB-600M), and decoder-only LLMs (Gemma~3 family, 1B--12B, and TranslateGemma) across training-data sizes, model scales, and in-domain, cross-domain, multilingual and multi-domain settings. To the best of our knowledge, this is the most extensive systematically documented multi-domain parallel data creation and benchmarking effort for low-resource MT. Our data and models
will be publicly released.}
\end{abstract}
\section{Introduction}

Solomonic ability shown by Large Language Models (LLMs) in solving language-related problems in the context of high-resource languages is not replicated when it comes to Machine Translation (MT) in low-resource languages (LRLs)~\cite{pang2025salute}. Although LLMs, just like their predecessors~\cite{lample2018word}, exhibit emergent cross-lingual capabilities~\cite{wang2024probing}, they are still constrained by limited representation of LRLs in pre-training corpora and insufficient exposure to parallel data.
% Although LLMs exhibit emergent cross-lingual capabilities, their MT performance is still constrained by limited representation of LRLs in pre-training corpora and insufficient exposure to parallel data. 
{This challenge is exacerbated in domain-specific contexts, where parallel corpora for LRLs are often sparse or entirely absent~\cite{ranathunga2024exploiting}. This highlights the need for continuous investing on parallel dataset creation efforts targetting LRLs.}
%This challenge is exacerbated by the limited availability of parallel corpora for LRLs in domain-specific contexts, where data is often sparse or entirely absent~\cite{ranathunga2024exploiting}. 
%\ndstodo{I feel the following sentence is a bit signpost-y, something that is frowned upon these days due to AI}Addressing this limitation is critical for real-world applications, especially in high-stakes scenarios such as pandemics, natural disasters, and other global crises, where timely and accurate multilingual communication is essential.

We present a parallel dataset that contains train/test data across seven domains (movie subtitles, Mathematics, Health, News, History (Wikipedia), Literature and open-domain), plus test data for one additional domain (LLM Jailbreak)  for three language pairs: English-Tamil, English-Sinhala and Sinhala-Tamil.  Parallel corpora, obtained from noisy web-mined corpora or by translation via  MT systems, were manually post-edited by professional translators to compile a training dataset of 200k+ sentences pairs, across the three language pairs. All the test data (10k+ sentences pairs) was manually created by the same translators. 

Using these datasets along with existing train/test data for two other domains (Government and Bible), as well as test data from FLORES+~\cite{nllb2024scaling}, we carry out an extensive set of experiments on domain-specific NMT across 11 test sets, for the three language pairs. Specifically, we evaluate the performance of a vanilla seq-seq translation model using Fairseq toolkit~\cite{ott-etal-2019-fairseq},  a pre-trained encoder-decoder translation model  (NLLB-600)~\cite{nllb2022} and an LLM (Gemma 3-1B instruct)~\cite{team2025gemma}. We further investigate how training data size differences, as well as model scaling, impact NMT results. We quantify the impact of divergence across domains when using existing data to build NMT systems for unseen domains, as well as the benefit of multi-domain, multilingual, as well as multi-domain + multilingual fine-tuning. 

To the best of our knowledge, this is the first study in LRL-MT to release per-domain training data and manually create test sets across this many domains at scale, together with systematic in-domain, cross-domain, multilingual and multi-domain experiments. \textbf{Our data and models will be publicly released.}

\section{Related Work}
\label{sec:lit}
\subsection{MT Datasets for LRLs}
There have been many NMT dataset creation efforts for LRLs in the past. These efforts can be broadly categorised as parallel corpus mining from the web~\cite{el2020ccaligned, banon2020paracrawl, schwenk2021ccmatrix}, manually correcting Machine Translated text~\cite{bhattacharjee2025coril}, or full manual translation~\cite{winata2023nusax, singh2025leveraging}. 

However, most of them focused on a single domain, or there is no specific domain. Some research that did create multi-domain parallel data did not carry out any domain-related MT experiments~\cite{premjith2019neural}. An exception is~\citet{bhattacharjee2025coril}. They created a dataset that covers three domains: Government, health and general, across 11 Indic languages. They carried out domain-specific experiments, however, only considered in-domain experiments (testing with a domain for which the NMT model was trained). Similarly,~\citet{appicharla2026maithilimt} presented multi-domain experiments across four domains for Hindi-Mailithi and~\citet{singh2025leveraging} presented a dataset across three domains for Bhili-Hindi. Experiments with these datasets have been limited to training and testing with a single domain at a time.  {Other closest resources to ours are MENYO-20k~\cite{adelani-etal-2021-effect}, an English--Yor\`ub\'a corpus with six test domains used for domain-adaptation experiments, and IN22~\cite{gala2023indictrans2}, human-translated test data over 13 domains for 22 Indic languages ($\sim$80 sentences per domain). However, neither releases per-domain training data at scale, nor supports cross-domain experiments.}

\subsection{Domain-specific NMT Evaluation}
Some other research focused on evaluating various facets of domain-specific NMT for LRLs, using existing data~\cite{ranathunga2024exploiting, khiu2024predicting, nayak2023leveraging}. Most of these used fine-tuned encoder-decoder pre-trained language models such as mBART~\cite{tang2021multilingual}. Out of these,~\citet{ranathunga2024exploiting} is the most comprehensive - they investigated how fine-tuned mBART performs across different domains, as well as how domain divergence impacts cross-domain performance. However, all these studies considered less than 5 domains. %TODO: write about HRL papers.

\subsection{Sinhala-Tamil-English MT}
Sinhala\footnote{Sinhala is also referred to as \textit{Sinhalese}, \textit{Singhala}, and \textit{Singhalese}~\cite{englebretson2005santa}.} is the native language of the Sinhalese people (approximately 17 million) and the only official language unique to Sri Lanka~\aliascite{parliament2022constitution}. Sinhala belongs to the Indo-European language family and the Indo-Aryan branch~\cite{arangala2024location}. Sinhala employs its own writing system and is morphologically rich~\cite{sirisoma1990brahmi}.

Tamil is the other official language in Sri Lanka. Tamil belongs to the Dravidian language family~\cite{krishnamurti2003dravidian}. It employs its own Brahmic script~\cite{daniels1996world} and is  an agglutinative language with rich morphology. Although Tamil has a more mature NLP ecosystem than Sinhala~\cite{de2026survey}, it remains mid-resourced~\cite{ranathunga-de-silva-2022-languages}. In particular, publicly available resources for Sri Lankan Tamil remain limited, and dialectal variation continues to hinder robust NLP applications~\cite{mahaganapathy-etal-2026-bridging}.

According to~\citet{ranathunga2022some}'s categorisation, Sinhala is classified as a low-resource language, and Tamil as a mid-resource language.  

Being a multilingual country, MT systems are of paramount importance to Sri Lanka, especially in domains such as education, health and government communications. The dearth of qualified human translators further reinforces the need for MT systems~\cite{farhath2018integration}. 

As a result, there have been research efforts to build MT systems for these language pairs (specifically Si-Ta), for more than a decade.~\citet{pushpananda2014sinhala} trained the first Sinhala-Tamil SMT systems on 25k parallel sentences, followed by several data augmentation efforts to improve the same~\cite{farhath2018improving,fernando2020data, farhath2018integration}. Subsequent research has moved to Vanilla NMT~\cite{tennage2017neural,pramodya2020comparison}, with techniques such as Byte Pair Encoding~\cite{nissanka2020exploring}, back-translation~\cite{epaliyana2021improving}, data augmentation~\cite{fernando2021data} and transliteration~\cite{tennage2018transliteration} techniques improving its performance.~\citet{thillainathan2021fine, lee2022pre} experimented with fine-tuning pre-trained seq-seq language models such as mBART, which showed clear advantage over vanilla NMT.~\citet{ranathunga-etal-2024-quality} ultimately showed that fine-tuning the translation-specific NLLB model~\cite{nllb2022} outperforms a fine-tuned mBART model.

None of this research publicly released the MT corpora it developed. Several publicly available web-mined corpora, such as CCMatrix~\citep{schwenk2021ccmatrix}, CCAlign~\citep{el2020ccaligned}, WikiMatrix~\citep{schwenk2021wikimatrix} and NLLB~\citep{nllb2022} cover the languages investigated in this study. However, as reported by~\citet{fernando-etal-2025-improving,ranathunga-etal-2024-quality}, such corpora are extremely noisy. FLORES and MultiMWP corpus~\cite{gamage2025multilingual} are the only human-curated publicly available corpora that cover all three language pairs, according to our knowledge. The former is an open-domain corpus. The latter only contains Math word problems.

\section{Multi-Domain Dataset Preparation}

\subsection{Data Sources and Preparation}
We constructed parallel datasets from two types of sources: publicly available parallel corpora and domain-specific monolingual corpora. Table~\ref{tab:datasources} shows data source details. Data selection was constrained by available data sources, translators, as well as funding.

\paragraph{Publicly Available Parallel Corpora: } As mentioned in Section~\ref{sec:lit}, publicly available web-mined corpora are extremely noisy. Therefore, following~\citet{ranathunga-etal-2024-quality}, we applied sentence-level filtering to the \NLLB{} and \OpenSubtitles{} corpora. Specifically, we deduplicated each corpus based on the English side and scored each source–target sentence pair using LASER-3~\cite{heffernan2022bitext}. Then, from the ranked corpus, we selected the top 30,000 sentence pairs, excluding sentences containing fewer than three words. We also used the Sinhala-Tamil-English portion of the MultiMWP corpus {(707 tri-aligned pairs and 439 EnTa pairs)}~\cite{gamage2025multilingual} as a training corpus. Unlike the aforementioned corpora, this is a clean, human curated dataset.

%\textcolor{red}{remove if we need to save space}During this process, we observed that a considerable portion of the highest-scoring sentence pairs in NLLB contained Bible-related content. Since we treat NLLB as an open-domain corpus, we removed such sentences by training a classifier and subsequently using a keyword-based filter. However, any remaining Bible-related examples that were only identified during manual inspection were retained, as they were considered part of the broader open-domain distribution. The filtered corpora were then corrected by human translators, resulting in 25,000 high-quality sentence pairs from each corpus.

\paragraph{Monolingual Corpora for Machine Translation: } We obtained monolingual corpora from sources where the data can be reused without copyright restrictions. Whenever possible, we consider documents related to Sri Lankan context. In most cases, monolingual data was in English; these corpora were translated into Sinhala and Tamil using Google Translate and IndicTrans2~\cite{gala2023indictrans2}, respectively. For part of the news corpus originally available in Sinhala, Sinhala sentences were first translated into English using Google Translate, and the post-edited English sentences were then translated into Tamil.

All web-mined and machine-translated sentences were post-edited by professional human translators to ensure translation quality.

\subsection{Test Set Creation}

For each of the training datasets, we created a test set entirely through manual translation. An additional test set was created using several corpora relating to LLM \Jailbreak{} prompts. Depending on translator availability, the validation set was created either through manual translation or by post-editing machine-translated sentences. Explicit checks were conducted to avaoid any overlapping between the three data splits. 

In addition, we used three existing parallel corpora in our experiments. First, we extracted 25,000 parallel verses from the \Bible{} corpus, using the data sources and scripts reported by~\citet{nayak2023leveraging}. The corresponding test and validation sets were extracted separately to ensure that there was no overlap with the training data. Second, we used 25,000 sentence pairs from a proprietary \Government-domain corpus ~\cite{fernando2021dataaugmentationterminologyintegration, ranathunga2018si}. For this corpus, we used the separately released validation and test sets provided by the original authors.

\begin{table*}[h]
\centering
\renewcommand{\arraystretch}{1.8}
\small
\begin{tabularx}{0.9\textwidth}{llX}
\toprule
\textbf{Corpus} &\textbf{Data Type} &\textbf{Sources} \\
\midrule
\NLLB & Parallel & \opus{https://opus.nlpl.eu/datasets/NLLB}{NLLB}
\\
\OpenSubtitles & Parallel &
\opus{https://opus.nlpl.eu/datasets/OpenSubtitles}{OpenSubtitles}\\

\Wikipedia & Monolingual & \makecell[l]{Local/Asian/European History - Pages before 1945\\
\url{https://www.wikipedia.org/}}\\

\News & Monolingual & \makecell[l]{Local News - \url{https://esana.com.lk/}\\
World News - \url{https://www.bbc.com/news/world}}\\

\Literature & Monolingual & \makecell[l]{\hf{https://huggingface.co/datasets/sil-ai/bloom-lm}{sil-ai/bloom-lm} \\
\url{https://www.gutenberg.org/}}\\

\Maths & Monolingual & \makecell[l]{\hf{https://huggingface.co/datasets/microsoft/orca-math-word-problems-200k}{microsoft/orca-math-word-problems-200k} \\
\url{https://www.gutenberg.org/}}\\

\Health & Monolingual
& \makecell[l]{\url{https://www.who.int/mega-menu/data/reports}\\
\url{https://fhb.health.gov.lk/resources}}\\

\Jailbreak &Monolingual & 
\makecell[l]{\citet{Liu2023JailbreakingCV}\\
\hf{https://huggingface.co/datasets/jackhhao/jailbreak-classification/blob/main/default/jailbreak_dataset_train.csv}{jackhhao/jailbreak-classification}}\\ 

% \FloresPlus &Parallel & \url{https://huggingface.co/datasets/openlanguagedata/flores_plus/tree/main/devtest}\\ 

\bottomrule
\end{tabularx}
\caption{The sources from which the parallel and monolingual data were obtained for corpus preparation}\label{tab:datasources}
\end{table*}

\subsection{Post-Editing \& Review Process}
Manual translation and post-editing were subject to a rigorous quality-control process. The translators for the task were selected via a screening process. Whenever possible, domain-specific translation experience was given priority. For example, health professionals/students with translation experience were selected for the health domain. %We publicly advertised for translators, and based on their professional qualifications and translation experience, they were shortlisted. Subsequently, a trial task was given, consisting of representative sentences sampled from the datasets used in this study. 
Selected translators were introduced to the task through a recorded demonstration video. In addition, we prepared customised translation guidelines for each domain and language pair; Appendix~\ref{app:training_materials} provides a sample guideline prepared for English–Sinhala translators. The well-experienced translators identified through the selection process were employed as reviewers.

%We initially recruited professional translators with prior translation experience and/or formal qualifications in translation or a related field. Final selection was based on their performance on a trial task consisting of representative sentences sampled from the datasets used in this study. 

For each domain, translators first completed post-editing tasks and subsequently carried out manual translation of the test sets. Translators' work was continuously reviewed and necessary feedback was provided. %Before assigning a full batch, each translator was given an initial set of 100 sentences to translate or post-edit. The output was reviewed by a more experienced translator. If more than 15\% of the sentences contained errors, the batch was returned with detailed feedback and additional instructions. Translators were assigned full batches of 1,000 sentences only if the quality of their initial 100-sentence batch was satisfactory; otherwise, their participation in the project was discontinued. For each subsequent 1,000-sentence batch, quality was checked by reviewing a sample of approximately 100 sentences.

This process was applied to both English–Sinhala and English–Tamil translation. To construct the Sinhala–Tamil corpus, we aligned the corresponding Sinhala and Tamil corpora using English as a pivot language.  If Sinhala–Tamil translators were available\footnote{Translators are much scarce for this pair.}, the parallel data underwent a trilingual alignment with further refinement.

%A summary of translator information is provided in Table \textcolor{red}{Aloka-Xxx} in Appendix~\ref{app:translator_details}. 
Over a period of three years, more than a 110 translators across the three language pairs contributed to the project. Translators were compensated in one of two ways, depending on their preference. Most translators received monetary compensation based on the government-approved translation rates of the relevant country. Others were offered co-authorship, following the increasingly adopted participatory research approach in NLP~\cite{chang2025global}, where contributors involved in language-data creation are recognised as research collaborators.

\subsection{Dataset Statistics \& Analysis}

Table~\ref{tab:corporastats} shows the statistics of the corpora we used for experiments. Not shown in this table are the 1000 sentence pairs per domain that we maintain as a hidden set, to be used in a future shared task. As mentioned earlier, \Bible, \Government, and \FloresPlus{} corpora are borrowed from prior research. The corpus we compiled totals up to 200k+ for training, 8k+ for validation and 11k+ for testing. Altogether, the corpus spans across 10 domains\footnote{The notion of a `domain' in MT is rather ambiguous. It can be interpreted with respect to topic and genre~\cite{van2015s}. However, identifying the exact domain of our datasets is out of scope for our work. Rather, we adopt~\citet{koehn2017six}'s definition of a domain in MT - `a domain is defined by a corpus from a specific source, and may differ from other domains in topic, genre, style, level of formality, etc.'}. 

\begin{table}[t]
  \centering
  %\small
  \footnotesize
  % \setlength{\tabcolsep}{4.3pt}
  % \resizebox{\linewidth}{!}{%
  % \resizebox{\dimexpr\columnwidth-2\tabcolsep\relax}{!}{%
 
    \begin{tabular}{@{\hspace{2pt}}c l @{\hspace{3.5pt}}l @{\hspace{6pt}}r r r}
      \toprule
      \# & Domain & Dir. & Train & Dev & Test \\
      \midrule

      % 3. Health
      \multirow{3}{*}{3} & \HealthC                          & \HealthC en-si    & \HealthC 1,000 & \HealthC 406 &    \HealthC 890 \\
                         & \HealthC                          & \HealthC en-ta    & \HealthC 2,000 & \HealthC 630 &  \HealthC 1,000 \\
                         & \multirow{-3}{*}{\HealthC Health} & \HealthC en-si-ta &    \HealthC -- &  \HealthC -- & \HealthC 784\dg \\
      \addlinespace[2pt]

      % Literature
      \multirow{4}{*}{5} & \LiteratureC                              & \LiteratureC en-si    &  \LiteratureC 6,674 & \LiteratureC 1,000 & \LiteratureC 1,000 \\
                         & \LiteratureC                              & \LiteratureC en-ta    &  \LiteratureC 3,500 &   \LiteratureC 511 &   \LiteratureC 864 \\
                         & \LiteratureC                              & \LiteratureC si-ta    &     \LiteratureC -- &    \LiteratureC -- &   \LiteratureC 831\dg \\
                         & \multirow{-4}{*}{\LiteratureC Literature} & \LiteratureC en-si-ta & \LiteratureC 802\dg &    \LiteratureC -- &    \LiteratureC -- \\
      \addlinespace[2pt]

      % Maths
      % 707 from parallel corpus ensi (5,900 without)
      % 1,146 from parallel corpus enta (3,499 without)
      \multirow{3}{*}{3} & \MathsC                         & \MathsC en-si    & \MathsC 6,607 & \MathsC 634 &    \MathsC -- \\
                         & \MathsC                         & \MathsC en-ta    & \MathsC 4,645 & \MathsC 644 &    \MathsC -- \\
                         & \multirow{-3}{*}{\MathsC Maths} & \MathsC en-si-ta &    \MathsC 707\dg &  \MathsC -- & \MathsC 1,000 \\
      \addlinespace[2pt]

      % News
      \multirow{3}{*}{4} & \NewsC                        & \NewsC en-si    &    \NewsC 16,337 &  \NewsC 1,000 &  \NewsC 1,000 \\
                         & \NewsC                        & \NewsC en-ta    &    \NewsC 10,000 &    \NewsC 602 &    \NewsC 996 \\
                         & \multirow{-3}{*}{\NewsC News} & \NewsC en-si-ta & \NewsC 10,000\dg & \NewsC 317\dg & \NewsC 992\dg  \\
      \addlinespace[2pt]

      % NLLB
      5 & \NLLBC NLLB & \NLLBC en-si-ta & \NLLBC 25,037 & \NLLBC 1,000 & \NLLBC 1,000 \\
      \addlinespace[2pt]

   % OpenSubtitles
      \multirow{4}{*}{6} & \OpenSubtitlesC                                 & \OpenSubtitlesC en-si    & \OpenSubtitlesC 25,007 &    \OpenSubtitlesC -- &    \OpenSubtitlesC -- \\
                         & \OpenSubtitlesC                                 & \OpenSubtitlesC en-ta    & \OpenSubtitlesC 28,350 &    \OpenSubtitlesC -- &    \OpenSubtitlesC -- \\
                         & \OpenSubtitlesC                                 & \OpenSubtitlesC si-ta    & \OpenSubtitlesC 25,111 &    \OpenSubtitlesC -- &    \OpenSubtitlesC -- \\
                         & \multirow{-4}{*}{\OpenSubtitlesC OpenSubtitles} & \OpenSubtitlesC en-si-ta &     \OpenSubtitlesC -- & \OpenSubtitlesC 1,000 & \OpenSubtitlesC 1,000 \\
      \addlinespace[2pt]

      % Wikipedia
      \multirow{3}{*}{7} & \WikipediaC                             & \WikipediaC en-si    &    \WikipediaC 24,160 &      \WikipediaC 964 &    \WikipediaC -- \\
                          & \WikipediaC                             & \WikipediaC en-ta    &    \WikipediaC 23,879 &       \WikipediaC -- &    \WikipediaC -- \\
                          & \multirow{-3}{*}{\WikipediaC Wikipedia} & \WikipediaC en-si-ta & \WikipediaC 20,502\dg & \WikipediaC 1,000 & \WikipediaC 1,000 \\
      \addlinespace[2pt]

    % Jailbreak
      8 & \JailbreakC Jailbreak & \JailbreakC en-si-ta & \JailbreakC -- & \JailbreakC -- & \JailbreakC 1,000 \\
      \addlinespace[2pt]

      % Bible
      \multirow{2}{*}{9} & \BibleC                         & \BibleC en-si & \BibleC 25,000 & \BibleC 1,000 & \BibleC 1,000 \\
                         & \multirow{-2}{*}{\BibleC Bible} & \BibleC en-ta & \BibleC 25,000 & \BibleC 1,000 & \BibleC 1,000 \\
      \addlinespace[2pt]

      % Government
      10 & \GovernmentC Government & \GovernmentC en-si-ta & \GovernmentC 25,000 & \GovernmentC 1,000 & \GovernmentC 1,000 \\
      \addlinespace[2pt]

      % FlorusPlus
      11 & \FloresPlusC FLORES+ & \FloresPlusC en-si-ta & \FloresPlusC -- & \FloresPlusC -- & \FloresPlusC 1,012 \\
      \addlinespace[2pt]

      \midrule
      \multicolumn{3}{l}{\textbf{Total}} & \textbf{277,307} & \textbf{12,391} & \textbf{14,762} \\
      \bottomrule
    \end{tabular}%
  % }
  \caption{Statistics of corpora used for experiments. Last three have been borrowed
    from other sources. Totals exclude trilingual (en-si-ta) subsets marked
    $^\dagger$, which are derived from or largely overlapping with the bilingual pairs listed for the same domain
    and split}
  \label{tab:corporastats}
\end{table}

\begin{figure*}[!hbt]	
	\centering 
\includegraphics[width=\linewidth]{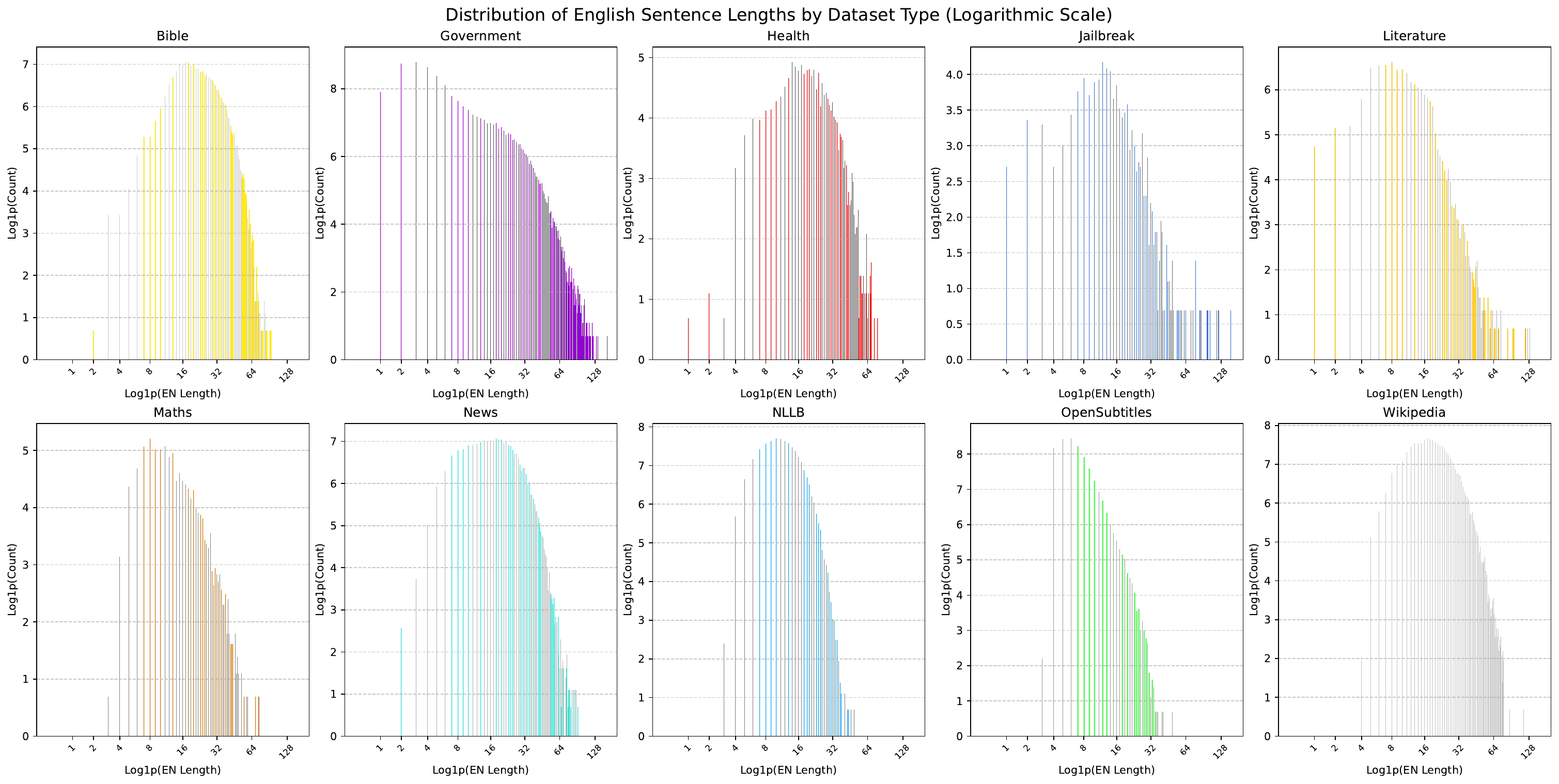}
\caption{Distribution of English Sentence Lengths by Domain (Logarithmic Scale)}
	\label{fig:EngDist}
\end{figure*}

{In Fig~\ref{fig:EngDist} we visualize the English sentence lengths of each domain using the same colour scheme as Table~\ref{tab:corporastats}. It can be noted that \Jailbreak{} dataset shows the most irregular distribution. This is unsurprising given the specialised purpose of that dataset. \Literature and \OpenSubtitles{} show similar distributions favouring shorter utterances; however, \Literature{} does have a considerable count of single-word exclamations or utterances as well as some rather long sentences. \Bible, \Government, and \Health{} show an affinity to longer sentences. \NLLB, \Bible, \Wikipedia, and \News{} show the closest resemblance to normal distributions.}

%\textcolor{red}{Nisansa - discuss figs 7,9,10}

{In order to further quantify the differences across domains, we calculated the Jensen–Shannon Divergence (JSD)~\cite{lin1991divergence}. Data sizes used for this calculation range from $\sim$16k to 25k (domains with limited training data were brought into this range by sourcing additional data considering their original distribution)\footnote{FLORES was not considered in this calculation, since we could not identify its exact source.}. Accordingly, the calculation considered only English data.} 

\begin{figure}[!htb]
    \centering
    \includegraphics[width=1\linewidth]{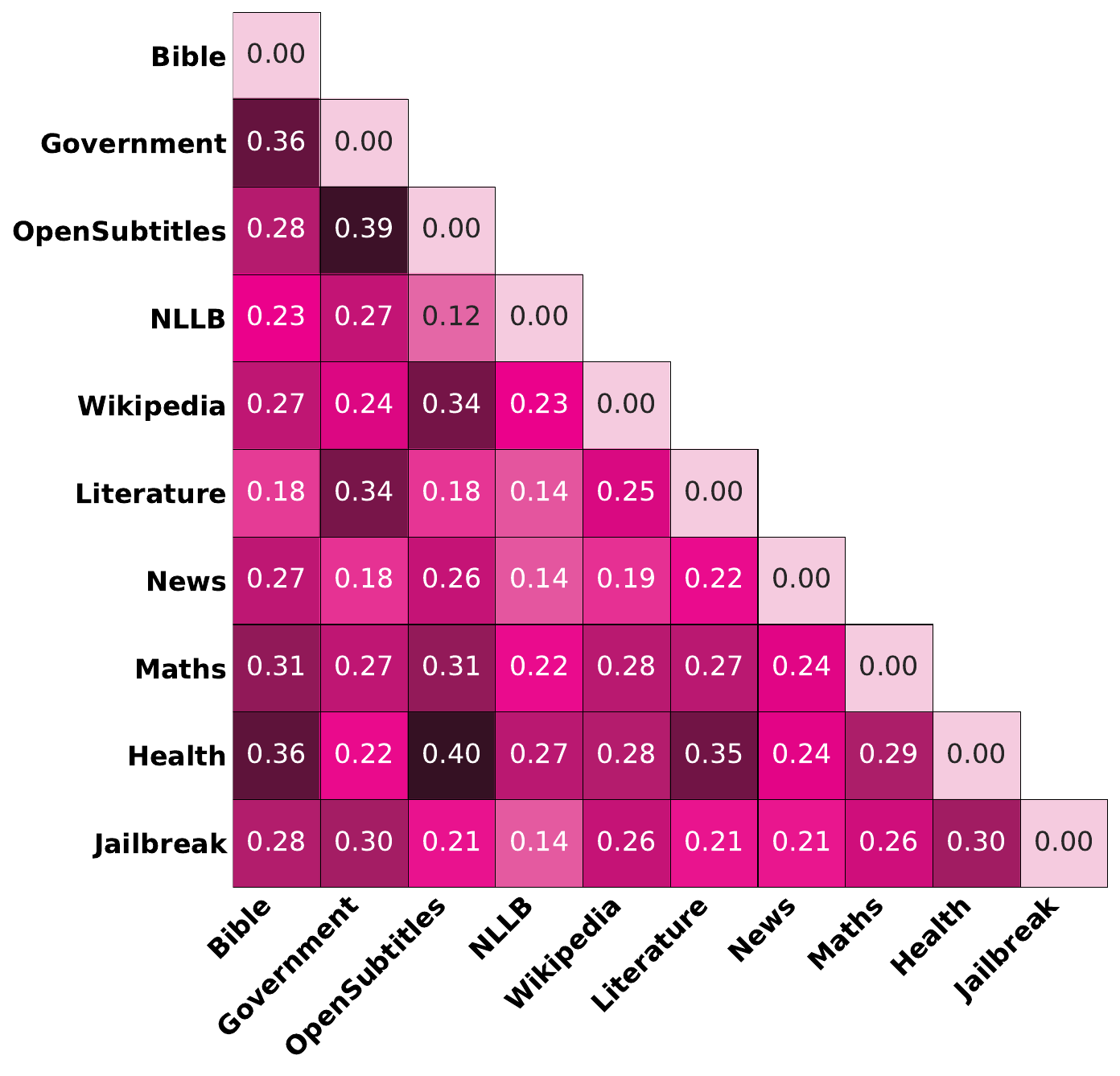}
    \caption{Pairwise JSD values between domains, computed on English training samples}
    \label{fig:JSD}
\end{figure}

Figure~\ref{fig:JSD} shows the pairwise JSD values between the 10 domains. Overall, \NLLB{} shows low divergence with \OpenSubtitles, \Literature, \News, and \Jailbreak, while \Literature{} also shows low divergence with \Bible{} and as does \News{} with \Government. These pairs have JSD values between 0.12 and 0.18. \NLLB, being open-domain, has the lowest divergence against four other domains, which could be due to it being a web-mined corpus. In contrast, \OpenSubtitles{} shows high divergence with \Health, \Government, and \Wikipedia; \Health{} also diverges strongly from \Bible{} and \Literature{}, while \Bible{} and \Government{} form another highly divergent pair. These pairs have JSD values between 0.34 and 0.40. \OpenSubtitles, which has shorter, spoken-style sentences, has the highest divergence against three other domains. Similarly, \Health, with its highly domain-specific terminology, also shows high divergence against three other domains.
%{Figure~\ref{fig:JSD} shows the Pairwise JSD values between the 10 domains. Overall, OpenSubtitles–NLLB, NLLB–Literature, NLLB–News, NLLB–Jailbreak, Bible–Literature and Government–News domain pairs show the lowest divergence among them, with JSD values between 0.12 and 0.18. NLLB, being open-domain has the lowest divergence against four other domains, which could be due to it being a web-mined corpus. On the other hand, OpenSubtitles–Health, Government–OpenSubtitles, Bible–Government, Bible–Health, Literature–Health and OpenSubtitles–Wikipedia pairs have the highest divergence, with JSD values between 0.34 and 0.40. \OpenSubtitles, which has shorter, spoken-style sentences has the highest divergence against three other domains. Similarly, \Health, with its highly domain-specific terminology also shows a high divergence against three other domains.}

\begin{figure*}[t!]
    \centering
    \begin{subfigure}[t]{0.495\linewidth}
        \centering
        \includegraphics[width=\linewidth]{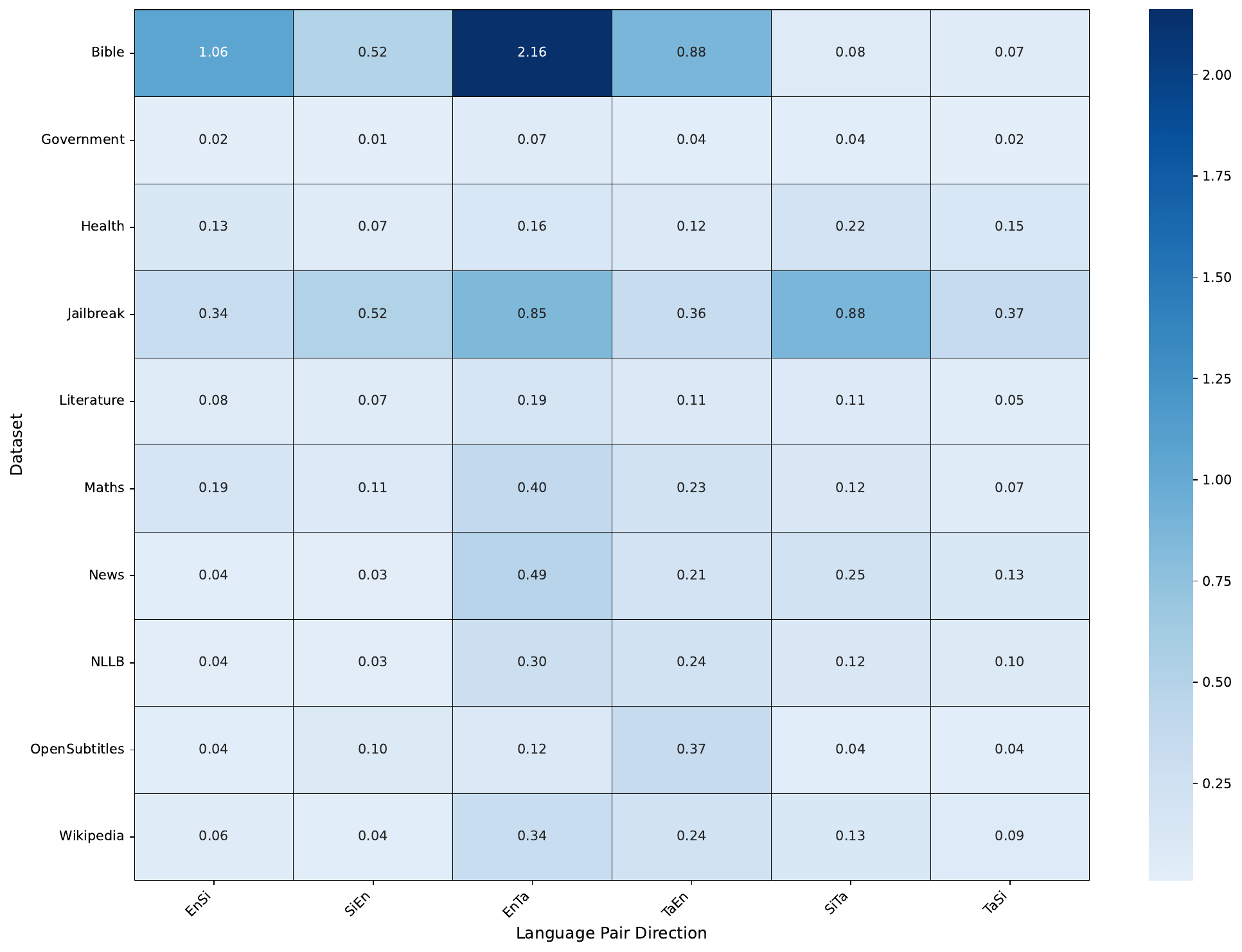}
        \caption{Sentence Length}
        \label{fig:KLSL}
    \end{subfigure}%
    \begin{subfigure}[t]{0.495\linewidth}
        \centering
        \includegraphics[width=\linewidth]{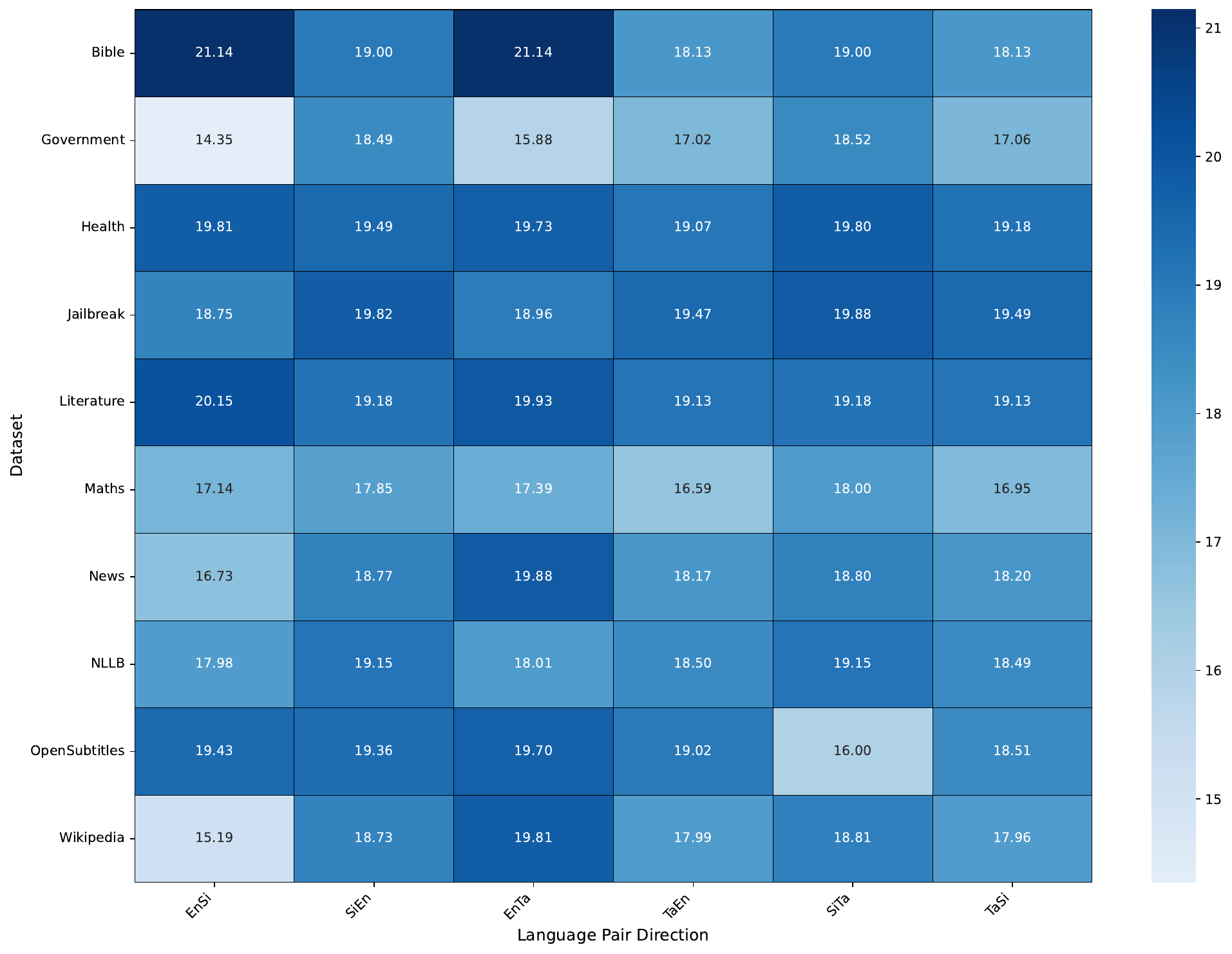}
        \caption{Word Frequencies}
        \label{fig:KLWF}
    \end{subfigure}
\caption{KL Divergence Across All Language Pair Directions by Dataset}
    \label{fig:KLSLWF}
\end{figure*}

{In Figure~\ref{fig:KLSLWF} we show the KL divergence across all language pair directions by dataset. The reason for using KLD for this analysis is twofold: (1) Unlike JSD, KLD is directional and thus allows us to show cases where a pair is relatively highly divergent in one direction and less so in the other. This is well evident in Figure~\ref{fig:KLSL} for \Bible, where we can observe high divergence in the EnSi and EnTa direction attesting the fact that Sinhala and Tamil languages had not evolved alongside Christianity as much as English did and as such, needs more words in a sentence to discuss something English would do so with fewer\footnote{One good example for this was the translation of the word "damned". Which is a simple enough concept to explain in a single word in English but in Sinhala, not only did it take 3 to 4 words, but also those phrases were not exactly the same across multiple occurrences.}, (2) JSD is bounded and thus high divergence saturate. In Figure~\ref{fig:KLWF}, we can see the intricacies of the the levels of divergence through KLD. But when the same visualization was created with JSD, all datasets got flattened out. For example, \Maths{} indiscriminately showed 0.62 across all language pairs while \NLLB, \News, \Wikipedia, and \OpenSubtitles all flattened out to 0.68. But with KLD in Figure~\ref{fig:KLWF} for example we can note the interesting observation of \Government{} having relatively low divergence in EnSi and EnTa direction owing to the fact that government domain words that did not exist in Sinhala or Tamil before the colonial rule were explicitly ``created'' usually as loanwords and at least as calques, resulting in lower divergence in word frequencies\footnote{It should be noted that these loanwords or calques are not always sourced from English. in fact, a considerable number of Sinhala words in this domain has roots in Portuguese or even Dutch who colonised Sri Lanka before the English. But this does not dilute the fact that one word in English in this domain can map to one word in Sinhala or Tamil, at a relatively high frequency.}.}

%\textcolor{red}{Nisansa - discuss figs 7,9,10}

\begin{figure*}[!htb]
    \centering
    \includegraphics[width=\textwidth]{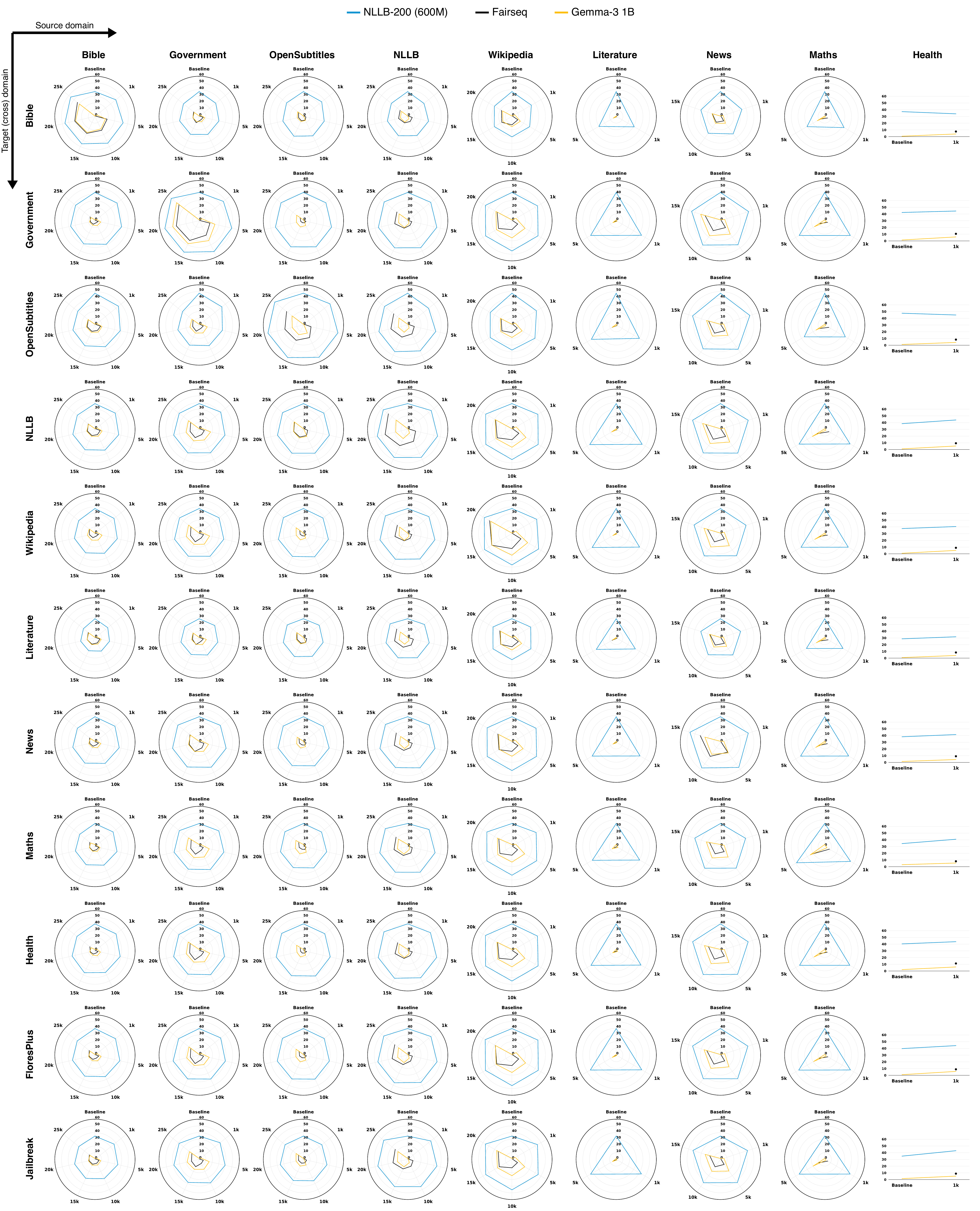}
    \caption{Source-domain (columns) vs cross-domain (rows) results for EnSi direction (chrF++)}
    \label{fig:grid1}
\end{figure*}

\begin{figure*}[!htb]
    \centering
    \includegraphics[width=\textwidth]{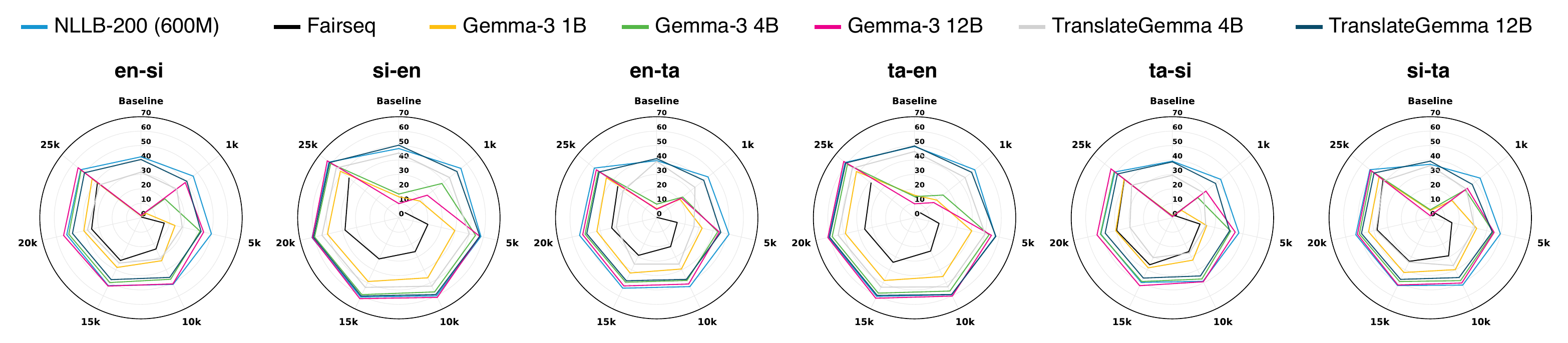}
    \caption{Language direction-wise results of  models trained and tested with Government data (chrF++)}
    \label{fig:grid2}
\end{figure*}

\begin{figure*}[!htb]
    \centering
    \includegraphics[width=\textwidth]{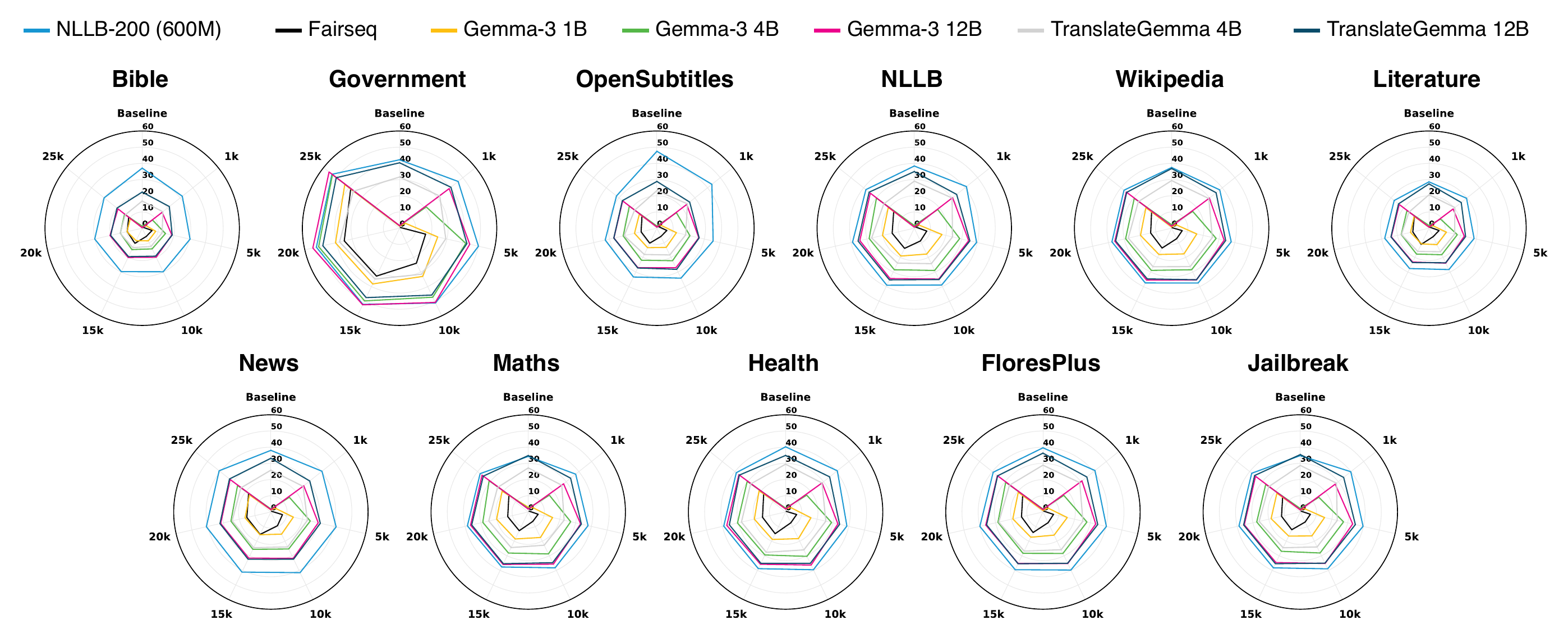}
    \caption {Cross-domain results of Government-trained models for EnSi direction (chrF++)}
    \label{fig:grid3}
\end{figure*}

% below needs gemma
\begin{figure*}[h]
    \centering
    \includegraphics[width=\textwidth]{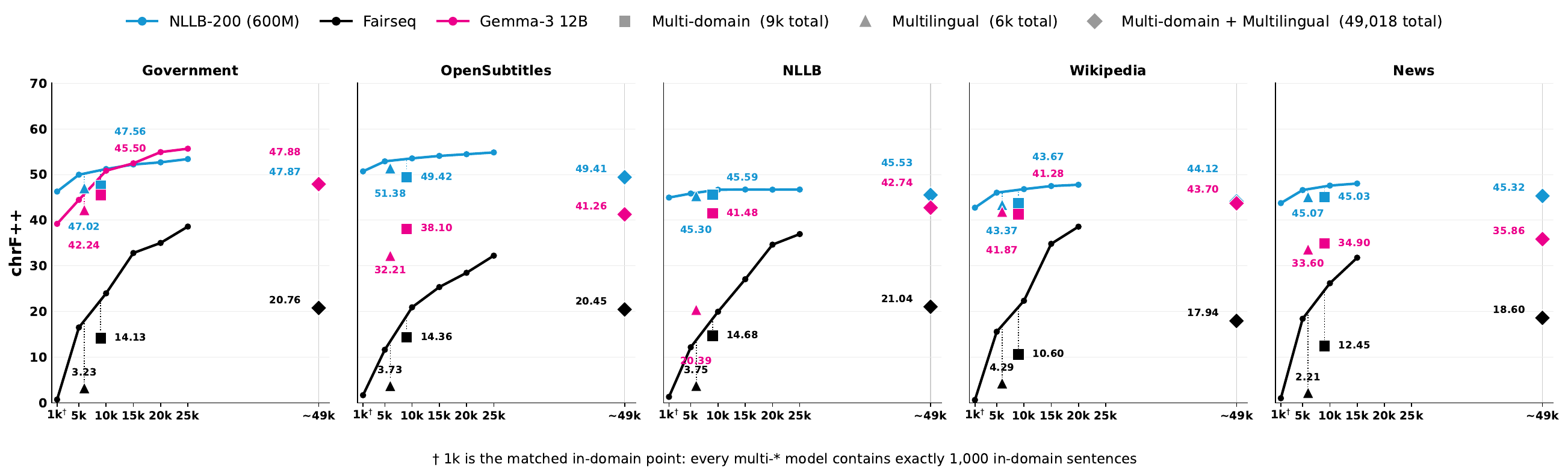}
    \caption{Strategy comparison across training data (chrF++) for EnSi}
    \label{fig:grid4}
\end{figure*}

\section{Experiments}
\subsection{Experiment Plan}
We conducted a comprehensive set of experiments to analyse the impact of training-data size, model architecture, and domain. We selected three types of models for fine-tuning: a standard encoder–decoder model implemented in Fairseq, a translation-specific encoder–decoder model (NLLB-600M)~\cite{nllb2022}, and an LLM %(Gemma 3.2 1B)\textcolor{red}{Isuru-instruct?}
{(the instruction-tuned Gemma~3 1B, \texttt{gemma-3-1b-it})}. Owing to computational resource constraints, we focused primarily on models with relatively small parameter counts. Each model was fine-tuned under %five training-data settings: 1k, 5k, 10k, 20k, and 25k sentence pairs.
{six training-data settings: 1k, 5k, 10k, 15k, 20k, and 25k sentence pairs.} We evaluated performance in both in-domain (models were trained and tested on data from the same domain) and out-of-domain (models were tested on a domain different to the training domain) settings.

We then conducted an ablation study to examine the effect of model choice. In addition to the models described above, we fine-tuned larger Gemma variants, namely {Gemma~3 4B and 12B, as well as TranslateGemma 4B and 12B}. To keep the experimental space tractable, this analysis was limited to the government domain. %This domain provides a challenging setting for LLM-based translation, as the data is proprietary and contains domain-specific terminology and style.

Third, we investigated the effects of multilingual fine-tuning, multi-domain fine-tuning, and combined multi-domain multilingual fine-tuning using Fairseq, Gemma 12B and NLLB. For this experiment, we sampled 1,000 sentence pairs from each domain for each language pair.%\kwtodo{explain the different setting for each setting (multilingual, multidomain, ml+md)}

%Finally, we assessed the impact of human post-editing. We fine-tuned NLLB-600M using the machine translated Wikipedia corpus, and compared its performance against model fine-tuned on the corresponding human post-edited corpus.\kwtodo{do we keep this?}

\subsection{Experiment Setup}%\kwtodo{add nllb/fairseq setup}
\label{sec:setup}

{We adapt all LLMs with Low-Rank Adaptation (LoRA)~\citep{hu2021lora}, using a minimal translation prompt in each model's native chat template. The recipe is held fixed across the Gemma~3 1B/4B/12B variants so that quality differences reflect model scale, whereas TranslateGemma uses a deliberately lighter recipe to limit erosion of its reinforcement-learned translation ability~\citep{finkelstein2026translategemma,biderman2024lora}, and is therefore assessed against its own zero-shot baseline rather than compared like-for-like with base Gemma~3. Prompt formats, full hyperparameters (Table~\ref{tab:hyperparams}), the rationale for the asymmetric recipes, and hardware details are given in Appendix~\ref{app:training_details}.}

\paragraph{Combined-corpus fine-tuning.}
%For the multi-domain experiment we fine-tune Gemma~3~12B on the concatenation of the
%training portions\textcolor{red}{Isuru-only 1k?} of all in-house domains, yielding one adapter per translation direction
%that is evaluated against each individual test domain; all other settings match the
%single-domain Gemma~3 recipe.\textcolor{red}{Isuru-how about other scenarios?}
%\textcolor{red}{Isuru-add prompt to appendix} -- prompt formats added to Appendix A
{The multi-domain experiment trains one adapter per translation direction on a mixture of 1k sentence pairs from each domain's training portion; the per-domain multilingual experiment trains, for each of the five domains with data in all three languages, one adapter jointly on all six translation directions; and the combined multilingual + multi-domain experiment trains a single adapter on all six directions of the multi-domain mixture. All three settings are run with NLLB, Fairseq, and Gemma~3~12B (the multi-domain setting also with TranslateGemma~12B), with all other settings matching the corresponding single-domain recipes.%; exact corpus sizes are given in Appendix~\ref{app:training_details}.}

\section{Automatic Evaluation }
Due to space limitations and brevity, we use ChrF++ for our discussion. Summary results for En-Si is shown in Figure~\ref{fig:grid1}. In particular, we noticed that COMET scores are inconsistent and unreliable. {For example, fully degenerate Gemma-1B outputs (chrF++ at or below 1.5) still receive COMET scores of 0.40 on average (up to 0.54)}, further affirming the observations in the prior literature~\cite{falcao2024comet}.% {All NLLB numbers in this section were updated after the earlier NLLB evaluations were found wrong and re-run (chrF++ only); non-NLLB numbers unchanged. The figures built on NLLB results (grid, row\_government\_langs, grid\_government\_multimodels, grid\_strategy\_ensi\_totaldata) need regeneration from the corrected chrF++ data. Also note the July multilingual/multi-domain numbers came from a stale grid: with corrected data the unified ML+MD adapter roughly ties the per-direction multi-domain adapters (NLLB 51.0 vs 50.9; Gemma-12B 50.6 vs 48.8), not +18-25 ahead.}

\subsection{Impact of training-data size, model architecture, and domain.}
\subsubsection{In-domain}
\paragraph{Individual Domain Complexity: }In-domain experiments with the vanilla Transformer model helps us understand the domain-wise translation difficulty. At 5k,  \Literature{} and \OpenSubtitles{} are consistently the hardest domains to learn from scratch; \Maths{} and \Bible{} the easiest.  \Literature{} and \OpenSubtitles, along with \NLLB{} are the hardest domains for Gemma-3. In contrast, for NLLB-600M, \OpenSubtitles{} becomes the easiest domain, while \Literature{} remains the hardest. This indicates the bias introduced during the pre-training stage. Even after training with 25k, the difference between the domains exists, with the maximum difference being about 10 ChRF++ points.  %\isuru{At 1k, the in-domain Fairseq results average 5.4 chrF++ with a standard deviation of 3.8 across the nine domains (language directions averaged first). The corresponding figures are 51.6 (std 4.6) for NLLB and 14.4 (std 7.0) for Gemma-1B. With more training data the domains converge for Fairseq and NLLB (on the four domains that reach 25k, the standard deviation falls to 2.8 and 3.8 respectively), but stay spread out for Gemma-1B (std 5.8 at 25k), whose web-mined domains ignite late. Literature is the hardest domain for all three models.}

In order to identify the reason behind this variation, we calculated the following per-domain statistics from the actual training/test corpora (English and target sides): average sentence length, vocabulary diversity (type–token ratio, i.e., distinct words per 10k running words), test-set OOV rate (share of test words never seen in the training data), and train–test overlap. We observe that the dominant factor for the Model training is the token budget, not lexical difficulty. Domains differ enormously in words per sentence. For example,  \OpenSubtitles{} has an average of 7.7 English words per sentence, \Literature{} 11.7, \Bible{} 25.2 and \Health{} has 26.1 words. This means, for example, the number of training tokens in 5k \Bible{} corpus is more than three times that of \OpenSubtitles. 

\paragraph{Model-Specific Performance: } In all the experiments, NLLB-600M always outperforms the other two models. However, its margin over Gemma-1B shrinks as the training dataset size increases. Gemma-1B outperforms Fairseq when the training dataset size is small, however the latter catches up, when the training dataset size rises to 25k.

\paragraph{Impact of Fine-tuning: }{NLLB is a strong zero-shot baseline on these domains (47.0 chrF++, averaged over the nine trainable domains and all language directions). Fine-tuning further improves the performance consistently: 1k in-domain pairs results in a gain of 4.6 ChRF++ points on average. When the dataset size goes to 25k, a further gain of 3.9 points is recorded, on average.  Gemma-1B starts near zero (8.0 average zero-shot), however it benefits immensely from fine-tuning, with a gain of 29.7 points over the zero-shot scores and gains up to +39.1 at full size, while Fairseq averages 5.4 at 1k and gains +7.3 to +40.6 from 1k to each domain's maximum. The three model classes therefore differ in where additional data helps: NLLB obtains about half of its total gain from the first 1k pairs, whereas Fairseq and Gemma-1B need volume before domain-specific data pays off. Interestingly, NLLB fine-tuned on just 1k pairs beats both competitors trained on each domain's entire corpus, in all 9 domains}

\subsubsection{Out-domain}
Out-domain results are mostly lower than the corresponding in-domain results, as expected (e.g. result for \Bible{} is less, when a model trained with another domain is used). Some domains seem to be more favourable than others. For example, for the  \Government{} domain, models trained with NLLB data are better than using \Bible{} data. This effect is more evident in the vanilla Transformer model. To investigate this further, we calculate the correlation between JSD and the result from each model. For the vanilla Transformer, an {r of -0.56 } suggests that there is a high negative correlation between the divergence between the two domains and the corresponding out-domain result. However, the impact of domain divergence becomes  less visible in the pre-trained models. %This observation aligns with that of the past research []. However, the impact of domain divergence becomes  less visible in the pre-trained models \isuru{(NLLB -0.19, Gemma-1B -0.23 at 25k)}. \isuru{The strength of this correlation also depends on the amount of training data: for the vanilla Transformer it is absent at 1k (+0.12, when its outputs are still degenerate) and strengthens monotonically with training size (-0.29 at 5k, -0.42 at 10k, -0.50 at 20k, -0.68 at 25k), whereas for the pretrained models it stays weak at every size. The divergence-transfer relationship is thus a property of models that must learn the language pair from the corpus at hand; pretraining removes most of the effect of domain divergence.}

We also note that fine-tuning NLLB with more data from a highly divergent domain steadily erodes its performance on the held-out domains. {Averaged over all language directions, its performance on FLORES+ drops by 7.2 chrF++ when the \Bible{} training set is increased from 1k to 25k, and by 2.1 chrF++ for \Government. In contrast, less divergent domains are free or even show a gain (open-web +0.8, \News{} +0.1, \OpenSubtitles{} +0.2, \Wikipedia{} +0.2 on FLORES+). The erosion rate correlates with the domain's mean divergence from the other domains (rank correlation -0.71 across the 8 domains). The \Jailbreak{} test set shows the same pattern (\Bible{} -9.2, \Government{} -2.3, broad domains near zero).} However, Gemma-1B does not seem to have this issue. Every domain improves its FLORES+ score {(+2.5 to +22.5)}. A model that is still learning to translate only benefits from more parallel data; forgetting is a phenomenon of already-competent translation models. {The same contrast appears within \texttt{TranslateGemma} under an identical light fine-tuning recipe: the 4B variant loses about 3 FLORES+ points while the 12B variant loses none; model scale also protects previously learned translation ability.}

%The open-domain web corpus is the best single "donor" for every model (transfers at 48.2 mean chrF++ for NLLB, vs Bible worst at 39.4). The largest in/out-domain penalty is on OpenSubtitles (-18.5 for NLLB, en→si), meaning conversational language is the hardest to reach from any other domain; Literature is the hardest transfer target overall.
{The best single ``donor'' domain is model-dependent: \Wikipedia{} for the vanilla Transformer (28.1 mean chrF++ across the other domains) and for Gemma-1B (33.8), and \Health{} for NLLB (50.4, from only 1k pairs); \Bible{} or \Literature{} is the worst donor for every model (NLLB with \Bible: 40.0). For NLLB the spread between donors is small, because its zero-shot performance is already strong. The largest in/out-domain penalty for NLLB is on \OpenSubtitles{} (in-domain 60.4 vs 56.9 for the best donor), so conversational language is the hardest register to reach from any other domain.}

\subsection{Language-wise Analysis}
Figures~\ref{fig:grid1} and~\ref{fig:grid2} depict these results. 
{Three independent lines of evidence identify generating Sinhala as the system-building bottleneck, beyond scores simply being lower. First, training collapses concentrate there: counting fine-tuned models whose output remains degenerate (chrF++ at or below 1.5), the vanilla Transformer has 159 such cases and Gemma-1B 73, every one of Gemma-1B's being into Sinhala (en-si 43, ta-si 30; none into Tamil or English), while NLLB has none anywhere. Into-Sinhala also needs roughly 5-10x more data to ignite for Gemma-1B (open-web corpus, en-si: 1.1 chrF++ at 5k, 6.2 at 10k, 22.3 at 25k). Second, a controlled comparison isolates the target side: the Sinhala-Tamil corpora contain the same sentence pairs in both directions, so si-ta vs ta-si compares generating Tamil vs Sinhala on identical content; si-ta wins in 17 of 19 model-domain cells, by up to +15.9 chrF++. Third, the gap survives fine-tuning for small models: at each domain's maximum size en-ta exceeds en-si by +11.6 chrF++ on average for NLLB and +15.8 for Gemma-1B, although the 4B/12B-class models close or reverse this gap on the one domain where we can test them (\Government: Gemma-4B -1.3, Gemma-12B -2.2, TranslateGemma-12B +0.9). Across all trained in-domain results the target-language averages are English 42.7, Tamil 37.9, Sinhala 31.3. The source side shows no comparable asymmetry: zero-shot NLLB's ta-en advantage over si-en is small and mixed in sign (-6.7 to +11.4 across domains), so the difficulty lies in generating Sinhala rather than in understanding it.}

\subsection{Effect of model choice}
As per Figures~\ref{fig:grid1} and~\ref{fig:grid2}, {on the \Government{} domain at 25k, the model ladder is Fairseq 40.0 < TranslateGemma-4B 42.3 < Gemma-1B 46.5 < TranslateGemma-12B 53.6 < Gemma-4B 54.6 < NLLB-600M 56.0 < Gemma-12B 57.1 (chrF++, averaged over all six directions). Base Gemma-3 scales cleanly with size once fine-tuned (46.5, 54.6, 57.1), yet a 600M-parameter translation-specific model beats every LLM except the 12B. Bigger is not better zero-shot: Gemma-12B is worse than Gemma-1B/4B zero-shot on the low-resource directions (e.g. si-en 7.6 vs 13.1/13.9), likely due to chat-style formatting and verbosity; fine-tuning restores the expected order, so zero-shot LLM quality is a poor predictor of post-fine-tuning quality at these scales. And TranslateGemma-12B, already translation-tuned, scores 43.2 zero-shot on \Government{} where same-size base Gemma-12B is near zero; fine-tuning adds a further +10.4 chrF++, while its 4B variant can degrade (en-ta 39.1 to 29.8), consistent with fine-tuning damaging an already-optimized translation policy at small scale.}

\subsection{Multilingual vs Multi-Domain}
We separately compiled a multi-domain and a multilingual corpus, by sampling out 1k from each domain and language pair. We trained the three models using these corpora separately, as well as by combining them.  According to Figure~\ref{fig:grid4}, if a given domain has a dedicated corpus of about 25k, it is generally better than having a multilingual+multi-domain corpus of double the size, where each domain is equally represented across language pairs. On the other hand, if a domain has a small corpus (less than 5k), combining it with data from other domains and language pairs is beneficial. Note that here we merged all the corpora, without considering the domain divergence. Hand-picking domains based on the domain divergence might result in better gains, however, exhaustive experiments to verify this is beyond the scope of the current research.

\section{Conclusion}

{We presented \EnSiTa, a trilingual multi-domain parallel dataset and
benchmark for English, Sinhala and Tamil, covering all six translation
directions and eight domains. Using this benchmark, we carried out an extensive
study of domain-specific MT for low-resource languages, spanning
training-data size, model architecture and scale, cross-domain transfer, and
multilingual and multi-domain fine-tuning. Our results translate into
concrete guidance for practitioners building domain-specific MT systems under
a limited budget: when starting from a pre-trained translation model, a small
curated in-domain set is on average worth more than any amount of
out-of-domain data, most strongly when translating into English,
whereas from-scratch models need volume before domain fit begins to
matter;
domain divergence reliably predicts transfer loss for from-scratch models but
matters far less once a model is pre-trained; and continued fine-tuning on a
divergent domain erodes the general translation ability of an
already-competent model, while a model still learning to translate only
benefits from more parallel data. We will release the dataset, the fine-tuned
models, and all evaluation outputs. We hope \EnSiTa{} enables the systematic study of
domain effects in low-resource MT that has so far been possible only for
high-resource languages, and we see document-level translation, additional
domains, and human evaluation of multi-domain systems as natural next
steps.}

\section*{Acknowledgements}
Data creation and curation was made possible by a Google Diversity and Inclusion grant received by Surangika Ranathunga and Nisansa de Silva. We thank  OpenToken (opentoken.global) for providing the GPU compute for LLM experiments, and LeafCloud (leaf.cloud) for the underlying infrastructure. We also  thank the National Languages Processing Centre (NLPC), at the  University of Moratuwa for providing the GPUs for Fairseq and NLLB-600 experiments. We also acknowledge the inputs from Madhavi Perera and Kengatharaiyar Sarveswaran during the initial phases of the project. We are grateful to Dhanika Perera from Bhasha Lanka (Pvt) Ltd for providing a Sinhala news corpus.

\bibliographystyle{acl_natbib}
% ---- Acknowledgments: omitted for anonymous submission (TACL rule). Restore this
% ---- section in the camera-ready, and restore the two named-partner mentions in
% ---- Appendix A (marked with "camera-ready:" comments there).
%\section*{Acknowledgments}
%All Gemma~3 and TranslateGemma fine-tuning and evaluation in this work was carried
%out on a persistent NVIDIA A100 80\,GB VM provided free of charge by OpenToken, with
%LeafCloud as the infrastructure partner. This compute grant was the enabling resource for
%the 4B/12B model-scaling and combined-corpus experiments, and we gratefully acknowledge
%their support. The NLLB-600M and Fairseq experiments were run on the internal GPU cluster
%of the University of Moratuwa, whose support we likewise gratefully acknowledge.
%\isurutodo{Camera-ready: restore the commented-out Acknowledgments section above and the named OpenToken/LeafCloud and University of Moratuwa mentions in Appendix~\ref{app:training_details}; confirm preferred crediting wording with OpenToken/LeafCloud.}

\bibliography{anthologyBib/tacl2021-Arxiv.bib}

@STRING(ACL = "Association for Computational Linguistics")

@article{pang2025salute,
author = "Pang, Jianhui and Ye, Fanghua and Wong, Derek Fai and Yu, Dian and Shi, Shuming and Tu, Zhaopeng and Wang, Longyue",
title = "Salute the classic: Revisiting challenges of machine translation in the age of large language models",
journal = "Transactions of the Association for Computational Linguistics",
volume = "13",
pages = "73--95",
year = "2025",
publisher = "MIT Press 255 Main Street, 9th Floor, Cambridge, Massachusetts 02142, USA\textasciitilde …"
}

@inproceedings{lample2018word,
author = "Lample, Guillaume and Conneau, Alexis and Ranzato, Marc'Aurelio and Denoyer, Ludovic and J{\'e}gou, Herv{\'e}",
title = "{Word Translation Without Parallel Data}",
booktitle = "International Conference on Learning Representations",
year = "2018"
}

@inproceedings{wang2024probing,
author = "Wang, Hetong and Minervini, Pasquale and Ponti, Edoardo",
title = "Probing the emergence of cross-lingual alignment during LLM training",
booktitle = "Findings of the Association for Computational Linguistics: ACL 2024",
pages = "12159--12173",
year = "2024"
}

@article{ranathunga2024exploiting,
author = "Ranathunga, Surangika and Nayak, Shravan and Lee, En-Shiun and Peng, Xin and Huang, Shih-Ting and Zeng, Yuchen and Mao, Yanke and Su, Tong and Chan, Yun-Hsiang and Yuan, Songchen and others",
title = "Exploiting domain-specific parallel data on multilingual language models for low-resource language translation",
journal = "ACM Transactions on Asian and Low-Resource Language Information Processing",
year = "2024",
publisher = "ACM New York, NY"
}

@article{nllb2024scaling,
author = "{NLLB Team} and Costa-jussà, Marta R. and Cross, James and Çelebi, Onur and Elbayad, Maha and Heafield, Kenneth and Heffernan, Kevin and Kalbassi, Elahe and Lam, Janice and Licht, Daniel and Maillard, Jean and Sun, Anna and Wang, Skyler and Wenzek, Guillaume and Youngblood, Al and Akula, Bapi and Barrault, Loic and Gonzalez, Gabriel Mejia and Hansanti, Prangthip and Hoffman, John and Jarrett, Semarley and Sadagopan, Kaushik Ram and Rowe, Dirk and Spruit, Shannon and Tran, Chau and Andrews, Pierre and Ayan, Necip Fazil and Bhosale, Shruti and Edunov, Sergey and Fan, Angela and Gao, Cynthia and Goswami, Vedanuj and Guzmán, Francisco and Koehn, Philipp and Mourachko, Alexandre and Ropers, Christophe and Saleem, Safiyyah and Schwenk, Holger and Wang, Jeff",
title = "{Scaling neural machine translation to 200 languages}",
journal = "Nature",
pages = "1--6",
year = "2024",
publisher = "Nature Publishing Group UK London"
}

@inproceedings{ott-etal-2019-fairseq,
author = "Ott, Myle and Edunov, Sergey and Baevski, Alexei and Fan, Angela and Gross, Sam and Ng, Nathan and Grangier, David and Auli, Michael",
title = "fairseq: A Fast, Extensible Toolkit for Sequence Modeling",
booktitle = "Proceedings of the 2019 Conference of the North {A}merican Chapter of the Association for Computational Linguistics (Demonstrations)",
year = "2019",
publisher = ACL,
doi = "10.18653/v1/N19-4009",
pages = "48--53"
}

@article{nllb2022,
author = "{NLLB Team} and Costa-jussà, Marta R. and Cross, James and Çelebi, Onur and Elbayad, Maha and Heafield, Kenneth and Heffernan, Kevin and Kalbassi, Elahe and Lam, Janice and Licht, Daniel and Maillard, Jean and Sun, Anna and Wang, Skyler and Wenzek, Guillaume and Youngblood, Al and Akula, Bapi and Barrault, Loic and Gonzalez, Gabriel Mejia and Hansanti, Prangthip and Hoffman, John and Jarrett, Semarley and Sadagopan, Kaushik Ram and Rowe, Dirk and Spruit, Shannon and Tran, Chau and Andrews, Pierre and Ayan, Necip Fazil and Bhosale, Shruti and Edunov, Sergey and Fan, Angela and Gao, Cynthia and Goswami, Vedanuj and Guzmán, Francisco and Koehn, Philipp and Mourachko, Alexandre and Ropers, Christophe and Saleem, Safiyyah and Schwenk, Holger and Wang, Jeff",
title = "No language left behind: Scaling human-centered machine translation",
journal = "arXiv preprint arXiv:2207.04672",
year = "2022"
}

@article{team2025gemma,
author = "{Gemma Team} and Kamath, Aishwarya and Ferret, Johan and Pathak, Shreya and Vieillard, Nino and Merhej, Ramona and Perrin, Sarah and Matejovicova, Tatiana and Ram{\'e}, Alexandre and Rivi{\`e}re, Morgane and others",
title = "Gemma 3 technical report",
journal = "arXiv preprint arXiv:2503.19786",
year = "2025"
}

@inproceedings{el2020ccaligned,
author = "El-Kishky, Ahmed and Chaudhary, Vishrav and Guzm{\'a}n, Francisco and Koehn, Philipp",
title = "CCAligned: A massive collection of cross-lingual web-document pairs",
booktitle = "Proceedings of the 2020 Conference on Empirical Methods in Natural Language Processing (EMNLP)",
pages = "5960--5969",
year = "2020"
}

@inproceedings{banon2020paracrawl,
author = "Ba{\\textasciitilde n}{\'o}n, Marta and Chen, Pinzhen and Haddow, Barry and Heafield, Kenneth and Hoang, Hieu and Espl{\`a}-Gomis, Miquel and Forcada, Mikel L and Kamran, Amir and Kirefu, Faheem and Koehn, Philipp and others",
title = "ParaCrawl: Web-scale acquisition of parallel corpora",
booktitle = "Proceedings of the 58th annual meeting of the association for computational linguistics",
pages = "4555--4567",
year = "2020"
}

@inproceedings{schwenk2021ccmatrix,
author = "Schwenk, Holger and Wenzek, Guillaume and Edunov, Sergey and Grave, Edouard and Joulin, Armand and Fan, Angela",
title = "CCMatrix: Mining billions of high-quality parallel sentences on the web",
booktitle = "Proceedings of the 59th Annual Meeting of the Association for Computational Linguistics and the 11th International Joint Conference on Natural Language Processing (Volume 1: Long Papers)",
pages = "6490--6500",
year = "2021"
}

@article{bhattacharjee2025coril,
author = "Bhattacharjee, Soham and Roy, Mukund K and Poojary, Yathish and Dave, Bhargav and Raj, Mihir and Mujadia, Vandan and Gain, Baban and Mishra, Pruthwik and Ahsan, Arafat and Krishnamurthy, Parameswari and others",
title = "CorIL: Towards Enriching Indian Language to Indian Language Parallel Corpora and Machine Translation Systems",
journal = "arXiv preprint arXiv:2509.19941",
year = "2025"
}

@inproceedings{winata2023nusax,
author = "Winata, Genta Indra and Aji, Alham Fikri and Cahyawijaya, Samuel and Mahendra, Rahmad and Koto, Fajri and Romadhony, Ade and Kurniawan, Kemal and Moeljadi, David and Prasojo, Radityo Eko and Fung, Pascale and others",
title = "NusaX: Multilingual parallel sentiment dataset for 10 Indonesian local languages",
booktitle = "Proceedings of the 17th Conference of the European Chapter of the Association for Computational Linguistics",
pages = "815--834",
year = "2023"
}

@article{singh2025leveraging,
author = "Singh, Pooja and Bhardwaj, Shashwat and Sharma, Vaibhav and Kumar, Sandeep",
title = "Leveraging the Cross-Domain \& Cross-Linguistic Corpus for Low Resource NMT: A Case Study On Bhili-Hindi-English Parallel Corpus",
journal = "arXiv preprint arXiv:2511.00486",
year = "2025"
}

@article{premjith2019neural,
author = "Premjith, B and Kumar, M Anand and Soman, KP",
title = "Neural machine translation system for English to Indian language translation using MTIL parallel corpus",
journal = "Journal of Intelligent Systems",
volume = "28",
number = "3",
pages = "387--398",
year = "2019",
publisher = "De Gruyter"
}

@article{appicharla2026maithilimt,
author = "Appicharla, Ramakrishna and Jha, Saroj Kumar and Ekbal, Asif and Bhattacharyya, Pushpak",
title = "Maithilimt: Developing Multi-Domain Parallel Corpus for Hindi-Maithili Machine Translation",
journal = "Language Resources and Evaluation",
volume = "60",
number = "1",
pages = "12",
year = "2026",
publisher = "Springer"
}

@inproceedings{adelani-etal-2021-effect,
author = "Adelani, David Ifeoluwa and Ruiter, Dana and Alabi, Jesujoba O. and Adebonojo, Damilola and Ayeni, Adesina and Adeyemi, Mofe and Awokoya, Ayodele Esther and Espa{\\textasciitilde n}a-Bonet, Cristina",
title = "The Effect of Domain and Diacritics in {Y}oruba{--}{E}nglish Neural Machine Translation",
booktitle = "Proceedings of Machine Translation Summit XVIII: Research Track",
year = "2021",
publisher = "Association for Machine Translation in the Americas",
url = "https://aclanthology.org/2021.mtsummit-research.6/",
pages = "61--75"
}

@article{gala2023indictrans2,
author = "Gala, Jay P and Chitale, Pranjal A and Gumma, Varun and Doddapaneni, Sumanth and Aswanth, Kumar M and Nawale, Janki Atul and Sujatha, Anupama and Puduppully, Ratish and Raghavan, Vivek and Kumar, Pratyush and others",
title = "IndicTrans2: Towards High-Quality and Accessible Machine Translation Models for all 22 Scheduled Indian Languages.",
journal = "Transactions on Machine Learning Research",
volume = "2023",
year = "2023"
}

@inproceedings{khiu2024predicting,
author = {Khiu, Eric and Toossi, Hasti and Anugraha, David and Liu, Jinyu and Li, Jiaxu and Flores, Juan and Roman, Leandro and Do{\u{g}}ru{\"o}z, A Seza and Lee, En-Shiun},
title = "Predicting machine translation performance on low-resource languages: The role of domain similarity",
booktitle = "Findings of the Association for Computational Linguistics: EACL 2024",
pages = "1474--1486",
year = "2024"
}

@article{nayak2023leveraging,
author = "Nayak, Shravan and Ranathunga, Surangika and Thillainathan, Sarubi and Hung, Rikki and Rinaldi, Anthony and Wang, Yining and Mackey, Jonah and Ho, Andrew and Lee, En-Shiun Annie",
title = "Leveraging auxiliary domain parallel data in intermediate task fine-tuning for low-resource translation",
journal = "arXiv preprint arXiv:2306.01382",
year = "2023"
}

@inproceedings{tang2021multilingual,
author = "Tang, Yuqing and Tran, Chau and Li, Xian and Chen, Peng-Jen and Goyal, Naman and Chaudhary, Vishrav and Gu, Jiatao and Fan, Angela",
title = "Multilingual translation from denoising pre-training",
booktitle = "Findings of the Association for Computational Linguistics: ACL-IJCNLP 2021",
pages = "3450--3466",
year = "2021"
}

@article{englebretson2005santa,
author = "Englebretson, R and Genetti, C",
title = "Santa Barbara papers in linguistics: Proceeding from the workshop on Sinhala linguistics",
journal = "Santa Barbara, CA: Department of Linguistics at the University of California, Santa Barbara",
year = "2005"
}

@book{parliament2022constitution,
author = "{Parliament of Democratic Socialist Republic of Sri Lanka}",
title = "The Constitution of The Democratic Socialist Republic of Sri Lanka",
url = "https://www.parliament.lk/files/pdf/constitution.pdf",
year = "2022",
publisher = "Parliament Secretariat"
}

@article{arangala2024location,
author = "Arangala, R",
title = "{Location of the Sinhala in Regional Linguistic Historicity and the Identity of Sinhala Language}",
journal = "Journal of Desk Research Review and Analysis",
volume = "2",
number = "1",
year = "2024"
}

@misc{sirisoma1990brahmi,
author = "Sirisoma, M. H.",
title = "{Brahmi inscriptions of Sri Lanka from 3rd century BC to 65 AD}",
journal = "Inscriptions: Volume Two, Archaeological Department Centenary (1890--1990), Commemorative Series",
pages = "3--54",
year = "1990",
publisher = "Department of Archaeology Colombo"
}

@book{krishnamurti2003dravidian,
author = "Krishnamurti, Bhadriraju",
title = "The dravidian languages",
year = "2003",
publisher = "Cambridge University Press"
}

@book{daniels1996world,
author = "Daniels, Peter T and Bright, William",
title = "The world's writing systems",
year = "1996",
publisher = "Oxford University Press on Demand"
}

@article{de2026survey,
author = "de Silva, Nisansa",
title = "{Survey on Publicly Available Sinhala Natural Language Processing Tools and Research}",
journal = "arXiv preprint arXiv:1906.02358v27",
year = "2026"
}

@inproceedings{ranathunga-de-silva-2022-languages,
author = "Ranathunga, Surangika and de Silva, Nisansa",
title = "Some Languages are More Equal than Others: Probing Deeper into the Linguistic Disparity in the {NLP} World",
booktitle = "Proceedings of the 2nd Conference of the Asia-Pacific Chapter of the Association for Computational Linguistics and the 12th International Joint Conference on Natural Language Processing (Volume 1: Long Papers)",
year = "2022",
publisher = ACL,
doi = "10.18653/v1/2022.aacl-main.62",
pages = "823--848"
}

@inproceedings{mahaganapathy-etal-2026-bridging,
author = "Mahaganapathy, Ahrane and Karunakaran, Sumirtha and Navakulan, Kavitha and Sarveswaran, Kengatharaiyer",
title = "Bridging Dialectal Variation: A Phonetic Transcription Tool for {T}amil",
booktitle = "Proceedings of the 13th Workshop on {NLP} for Similar Languages, Varieties and Dialects",
year = "2026",
publisher = ACL,
doi = "10.18653/v1/2026.vardial-1.19",
pages = "234--241"
}

@inproceedings{ranathunga2022some,
author = "Ranathunga, Surangika and De Silva, Nisansa",
title = "Some languages are more equal than others: Probing deeper into the linguistic disparity in the NLP world",
booktitle = "Proceedings of the 2nd Conference of the Asia-Pacific Chapter of the Association for Computational Linguistics and the 12th International Joint Conference on Natural Language Processing (Volume 1: Long Papers)",
pages = "823--848",
year = "2022"
}

@inproceedings{farhath2018integration,
author = "Farhath, Fathima and Ranathunga, Surangika and Jayasena, Sanath and Dias, Gihan",
title = "Integration of bilingual lists for domain-specific statistical machine translation for sinhala-tamil",
booktitle = "2018 Moratuwa Engineering Research Conference (MERCon)",
pages = "538--543",
year = "2018",
organization = "IEEE"
}

@inproceedings{pushpananda2014sinhala,
author = "Pushpananda, Randil and Weerasinghe, Ruvan and Niranjan, Mahesan",
title = "Sinhala-tamil machine translation: Towards better translation quality",
booktitle = "Proceedings of the Australasian Language Technology Association Workshop 2014",
pages = "129--133",
year = "2014"
}

@inproceedings{farhath2018improving,
author = "Farhath, Fathima and Theivendiram, Pranavan and Ranathunga, Surangika and Jayasena, Sanath and Dias, Gihan",
title = "Improving domain-specific SMT for low-resourced languages using data from different domains",
booktitle = "Proceedings of the Eleventh International Conference on Language Resources and Evaluation (LREC 2018)",
year = "2018"
}

@article{fernando2020data,
author = "Fernando, Aloka and Ranathunga, Surangika and Dias, Gihan",
title = "Data augmentation and terminology integration for domain-specific sinhala-english-tamil statistical machine translation",
journal = "arXiv preprint arXiv:2011.02821",
year = "2020"
}

@inproceedings{tennage2017neural,
author = "Tennage, Pasindu and Sandaruwan, Prabath and Thilakarathne, Malith and Herath, Achini and Ranathunga, Surangika and Jayasena, Sanath and Dias, Gihan",
title = "Neural machine translation for sinhala and tamil languages",
booktitle = "2017 International Conference on Asian Language Processing (IALP)",
pages = "189--192",
year = "2017",
organization = "IEEE"
}

@inproceedings{pramodya2020comparison,
author = "Pramodya, Ashmari and Pushpananda, Randil and Weerasinghe, Ruvan",
title = "A comparison of transformer, recurrent neural networks and SMT in Tamil to Sinhala MT",
booktitle = "2020 20th International Conference on Advances in ICT for Emerging Regions (ICTer)",
pages = "155--160",
year = "2020",
organization = "IEEE"
}

@inproceedings{nissanka2020exploring,
author = "Nissanka, LNASH and Pushpananda, BHR and Weerasinghe, AR",
title = "Exploring neural machine translation for sinhala-tamil languages pair",
booktitle = "2020 20th International Conference on Advances in ICT for Emerging Regions (ICTer)",
pages = "202--207",
year = "2020",
organization = "IEEE"
}

@inproceedings{epaliyana2021improving,
author = "Epaliyana, Koshiya and Ranathunga, Surangika and Jayasena, Sanath",
title = "Improving back-translation with iterative filtering and data selection for Sinhala-English NMT",
booktitle = "2021 Moratuwa Engineering Research Conference (MERCon)",
pages = "438--443",
year = "2021",
organization = "IEEE"
}

@inproceedings{fernando2021data,
author = "Fernando, Aloka and Ranathunga, Surangika",
title = "Data augmentation to address out of vocabularyproblem in low resource Sinhala English neural machine translation",
booktitle = "Proceedings of the 35th Pacific Asia Conference on Language, Information and Computation",
pages = "61--70",
year = "2021"
}

@inproceedings{tennage2018transliteration,
author = "Tennage, Pasindu and Herath, Achini and Thilakarathne, Malith and Sandaruwan, Prabath and Ranathunga, Surangika",
title = "Transliteration and byte pair encoding to improve tamil to sinhala neural machine translation",
booktitle = "2018 Moratuwa Engineering Research Conference (MERCon)",
pages = "390--395",
year = "2018",
organization = "IEEE"
}

@inproceedings{thillainathan2021fine,
author = "Thillainathan, Sarubi and Ranathunga, Surangika and Jayasena, Sanath",
title = "Fine-tuning self-supervised multilingual sequence-to-sequence models for extremely low-resource NMT",
booktitle = "2021 Moratuwa Engineering Research Conference (MERCon)",
pages = "432--437",
year = "2021",
organization = "IEEE"
}

@inproceedings{lee2022pre,
author = "Lee, En-Shiun Annie and Thillainathan, Sarubi and Nayak, Shravan and Ranathunga, Surangika and Adelani, David Ifeoluwa and Su, Ruisi and McCarthy, Arya D",
title = "Pre-trained multilingual sequence-to-sequence models: A hope for low-resource language translation?",
booktitle = "Findings of the Association for Computational Linguistics: ACL 2022",
pages = "58--67",
year = "2022"
}

@inproceedings{ranathunga-etal-2024-quality,
author = "Ranathunga, Surangika and de Silva, Nisansa and Velayuthan, Menan and Fernando, Aloka and Rathnayake, Charitha",
title = "Quality Does Matter: A Detailed Look at the Quality and Utility of Web-Mined Parallel Corpora",
booktitle = "Proceedings of the 18th Conference of the European Chapter of the Association for Computational Linguistics (Volume 1: Long Papers)",
year = "2024",
publisher = ACL,
doi = "10.18653/v1/2024.eacl-long.52",
pages = "860--880"
}

@inproceedings{schwenk2021wikimatrix,
author = "Schwenk, Holger and Chaudhary, Vishrav and Sun, Shuo and Gong, Hongyu and Guzm{\'a}n, Francisco",
title = "WikiMatrix: Mining 135M Parallel Sentences in 1620 Language Pairs from Wikipedia",
booktitle = "Proceedings of the 16th Conference of the European Chapter of the Association for Computational Linguistics: Main Volume",
pages = "1351--1361",
year = "2021"
}

@inproceedings{fernando-etal-2025-improving,
author = "Fernando, Aloka and de Silva, Nisansa and Velayuthan, Menan and Rathnayake, Charitha and Ranathunga, Surangika",
title = "Improving the Quality of Web-mined Parallel Corpora of Low-Resource Languages using Debiasing Heuristics",
booktitle = "Proceedings of the 2025 Conference on Empirical Methods in Natural Language Processing",
year = "2025",
publisher = ACL,
doi = "10.18653/v1/2025.emnlp-main.1435",
pages = "28264--28281",
ISBN = "979-8-89176-332-6"
}

@article{gamage2025multilingual,
author = "Gamage, Omega and Ranathunga, Surangika and Lee, Annie and Sun, Xiao and Singh, Aryaveer and Skenduli, Marjana Prifti and Alam, Mehreen and Nayak, Ajit Kumar and Gao, Haonan and Deori, Barga and others",
title = "A multilingual dataset (multimwp) and benchmark for math word problem generation",
journal = "IEEE Transactions on Audio, Speech and Language Processing",
volume = "33",
pages = "1838--1848",
year = "2025",
publisher = "IEEE"
}

@online{department2024Census,
author = "{Department of Census and Statistics, Sri Lanka}",
title = "{Census of Population and Housing of Sri Lanka}",
year = "2024",
url = "https://nisansa.lk/?s=AoRbxv",
lastaccessed = "24-July-2026"
}

@inproceedings{heffernan2022bitext,
author = "Heffernan, Kevin and {\c{C}}elebi, Onur and Schwenk, Holger",
title = "Bitext mining using distilled sentence representations for low-resource languages",
booktitle = "Findings of the Association for Computational Linguistics: EMNLP 2022",
pages = "2101--2112",
year = "2022"
}

@misc{fernando2021dataaugmentationterminologyintegration,
author = "Fernando, Aloka and Ranathunga, Surangika and Dias, Gihan",
title = "Data Augmentation and Terminology Integration for Domain-Specific Sinhala-English-Tamil Statistical Machine Translation",
year = "2021",
eprint = "2011.02821",
archivePrefix = "arXiv",
primaryClass = "cs.CL",
url = "https://arxiv.org/abs/2011.02821"
}

@inproceedings{ranathunga2018si,
author = "Ranathunga, Surangika and Farhath, Fathima and Thayasivam, Uthayasanker and Jayasena, Sanath and Dias, Gihan",
title = "Si-ta: Machine translation of sinhala and tamil official documents",
booktitle = "2018 National Information Technology Conference (NITC)",
pages = "1--6",
year = "2018",
organization = "IEEE"
}

@article{Liu2023JailbreakingCV,
author = "Liu, Yi and Deng, Gelei and Xu, Zhengzi and Li, Yuekang and Zheng, Yaowen and Zhang, Ying and Zhao, Lida and Zhang, Tianwei and Liu, Yang",
title = "Jailbreaking ChatGPT via Prompt Engineering: An Empirical Study",
journal = "ArXiv",
year = "2023",
volume = "abs/2305.13860",
url = "https://api.semanticscholar.org/CorpusID:258841501"
}

@article{chang2025global,
author = "Chang, Tyler A and Arnett, Catherine and Eldesokey, Abdelrahman and Sadallah, Abdelrahman and Kashar, Abeer and Daud, Abolade and Abosede, Grace Olanihun and Adamu, Labaran Mohammed and Praise, Adeyemi and Adhikarinayum, Meerajita Sharma and others",
title = "Global piqa: Evaluating physical commonsense reasoning across 100+ languages and cultures",
year = "2025"
}

@inproceedings{van2015s,
author = "Van der Wees, Marlies and Bisazza, Arianna and Weerkamp, Wouter and Monz, Christof",
title = "What’s in a domain? Analyzing genre and topic differences in statistical machine translation",
booktitle = "Proceedings of the 53rd Annual Meeting of the Association for Computational Linguistics and the 7th International Joint Conference on Natural Language Processing (Volume 2: Short Papers)",
pages = "560--566",
year = "2015"
}

@inproceedings{koehn2017six,
author = "Koehn, Philipp and Knowles, Rebecca",
title = "Six challenges for neural machine translation",
booktitle = "Proceedings of the first workshop on neural machine translation",
pages = "28--39",
year = "2017"
}

@article{lin1991divergence,
author = "Lin, Jianhua",
title = "Divergence measures based on the Shannon entropy",
journal = "IEEE Transactions on Information theory",
volume = "37",
number = "1",
pages = "145--151",
year = "1991",
publisher = "IEEE"
}

@inproceedings{hu2021lora,
author = "Hu, Edward J and Shen, Yelong and Wallis, Phillip and Allen-Zhu, Zeyuan and Li, Yuanzhi and Wang, Shean and Wang, Lu and Chen, Weizhu",
title = "Lo{RA}: Low-Rank Adaptation of Large Language Models",
booktitle = "International Conference on Learning Representations (ICLR)",
year = "2022"
}

@article{finkelstein2026translategemma,
author = "Finkelstein, Mara and Caswell, Isaac and Domhan, Tobias and Peter, Jan-Thorsten and Juraska, Juraj and Riley, Parker and Deutsch, Daniel and Kovacs, Geza and Dilanni, Cole and Cherry, Colin and Briakou, Eleftheria and Nielsen, Elizabeth and Luo, Jiaming and Black, Kat and Mullins, Ryan and Agrawal, Sweta and Xu, Wenda and Kats, Erin and Jaskiewicz, Stephane and Freitag, Markus and Vilar, David",
title = "Translate{G}emma Technical Report",
journal = "arXiv preprint arXiv:2601.09012",
year = "2026"
}

@article{biderman2024lora,
author = "Biderman, Dan and Portes, Jacob and Gonzalez Ortiz, Jose Javier and Paul, Mansheej and Greengard, Philip and Jennings, Connor and King, Daniel and Havens, Sam and Chiley, Vitaliy and Frankle, Jonathan and Blakeney, Cody and Cunningham, John P.",
title = "Lo{RA} Learns Less and Forgets Less",
journal = "Transactions on Machine Learning Research (TMLR)",
year = "2024"
}

@inproceedings{falcao2024comet,
author = "Falc{\\textasciitilde a}o, J{\'u}lia and Borg, Claudia and Aranberri, Nora and Abela, Kurt",
title = "COMET for low-resource machine translation evaluation: A case study of English-Maltese and Spanish-Basque",
booktitle = "Proceedings of the 2024 Joint International Conference on Computational Linguistics, Language Resources and Evaluation (LREC-COLING 2024)",
pages = "3553--3565",
year = "2024"
}

@inproceedings{dettmers2022optimizers,
author = "Dettmers, Tim and Lewis, Mike and Shleifer, Sam and Zettlemoyer, Luke",
title = "8-bit Optimizers via Block-wise Quantization",
booktitle = "International Conference on Learning Representations (ICLR)",
year = "2022"
}

@inproceedings{zhang2023prompting,
author = "Zhang, Biao and Haddow, Barry and Birch, Alexandra",
title = "Prompting Large Language Model for Machine Translation: A Case Study",
booktitle = "Proceedings of the 40th International Conference on Machine Learning (ICML)",
year = "2023"
}

\iftaclpubformat

\onecolumn

\appendix

\section{Training Configuration Details}\label{app:training_details}
Table~\ref{tab:hyperparams} lists the full LoRA fine-tuning hyperparameters for both model
families.

{In LoRA fine-tuning, the pre-trained weights are frozen and only low-rank adapters
injected into the attention and feed-forward projections are trained, using the Unsloth
library over Hugging Face Transformers and PEFT. For the multimodal
Gemma~3 checkpoints (4B/12B) the vision encoder is frozen and adapters are applied only to
the language-model layers. All training uses \texttt{bfloat16} with no weight quantization,
and the effective batch size is held constant across model sizes. The one choice essential
for stability was the 8-bit AdamW optimizer~\citep{dettmers2022optimizers}: 32-bit AdamW
produced NaN gradients early in training. Decoding is greedy (beam size $1$) with a budget
of 256 new tokens; spBLEU uses the FLORES-200 tokenizer and COMET the
\texttt{wmt22-comet-da} model. All fine-tuning and evaluation run on a single
NVIDIA~A100 80\,GB GPU, a persistent VM provided free of charge under a compute grant
% camera-ready: "provided by OpenToken with LeafCloud as the infrastructure partner"
(PyTorch~2.10, CUDA~12.8, Transformers~4.57, Unsloth, bitsandbytes),
with TF32 and the SDPA attention backend enabled; the full campaign (355 LoRA adapters and
on the order of 5,000 translate-and-score evaluations) used 720 A100 GPU-hours in total, as
measured by provider telemetry, including failed and exploratory runs.}

{The NLLB-600M and Fairseq experiments were run separately, on an institutional GPU
cluster
% camera-ready: "an internal GPU cluster at the University of Moratuwa"
%(\textcolor{red}{kavindu-GPU model and count, VRAM, software stack, and approximate GPU-hours}).}

{\paragraph{Prompt format.} For the base Gemma~3 models, each training example is
rendered in the Gemma chat template as a single user turn followed by the model turn:}

{\begin{quote}\small\ttfamily
<start\_of\_turn>user\\
\{SourceLanguage\}: \{source text\}\\
\{TargetLanguage\}:<end\_of\_turn>\\
<start\_of\_turn>model\\
\{target text\}<end\_of\_turn>
\end{quote}}

{where the language names are written out in English (English, Sinhala, Tamil) and the
loss is masked so that only the target-text tokens contribute to the
gradient~\citep{zhang2023prompting}. At inference
the prompt ends after \texttt{<start\_of\_turn>model} and the continuation is decoded
greedily. TranslateGemma does not use this string prompt: it keeps its native structured
translation message, a single user turn carrying \texttt{source\_lang\_code} and
\texttt{target\_lang\_code} fields (ISO~639-1 codes: en, si, ta) together with the source
text, rendered by its own chat template with the generation prompt appended, since
overwriting its instruction- and reinforcement-tuned prompt format degrades quality. In both
families, training and evaluation use identical prompts.}

{\paragraph{Combined-corpus training sizes.} The multi-domain mixture samples 1k
sentence pairs from each domain's training portion, giving 9,000 pairs for each
English-Sinhala and English-Tamil direction and 6,509 for Sinhala-Tamil, where only five
domains (\Government, \News, open-web, \OpenSubtitles, \Wikipedia) have data. The per-domain
multilingual adapters are trained on those same five domains, with 1k pairs per direction
(6,000 per domain); the combined multilingual + multi-domain adapter is trained on all six
directions of the multi-domain mixture, 49,018 examples in total.}

The two families are studied under \emph{different} research questions, namely the effect of
model scale for Gemma~3 versus the cost of domain-adapting an already-translation-tuned
model for TranslateGemma, so their recipes are intentionally not matched. Unlike the base
Gemma~3 checkpoints, TranslateGemma has already undergone supervised fine-tuning followed by
reinforcement learning that optimizes translation quality against an ensemble of automatic
reward models, including MetricX-QE and AutoMQM~\citep{finkelstein2026translategemma}.
Adapting it as aggressively as a general-purpose model would overwrite this
reinforcement-learned policy and induce catastrophic forgetting of precisely the translation
competence that makes the checkpoint valuable. Reducing the adapter rank, learning rate, and
number of epochs (and adding a small amount of dropout) is an established,
capability-preserving domain-adaptation strategy~\citep{biderman2024lora}: low-rank, smaller
updates constrain how far the adapted policy can drift from the pre-trained one, bounding the
erosion of general translation ability so that the resulting trade-off (in-domain gain versus
loss of general ability) can be \emph{measured} rather than masked. We consequently do not
claim a controlled, like-for-like comparison \emph{between} the two families: a difference in
score could reflect the recipe (rank, learning rate, epochs) as much as the base model.
Instead, each TranslateGemma adapter is assessed against its own zero-shot baseline, with
FLORES held out as an out-of-domain probe of general translation ability. This is a within-model
comparison that uses the same model and recipe on both sides and is thus unaffected by the
recipe asymmetry.

\begin{table}[t]
  \centering
  \small
  \begin{tabular}{lcc}
    \toprule
                         & Gemma~3        & Translate \\
                         & (1B/4B/12B)    & Gemma (4B/12B) \\
    \midrule
    LoRA rank $r$        & 32             & 16 \\
    LoRA $\alpha$        & 32             & 16 \\
    LoRA dropout         & 0.0            & 0.05 \\
    Adapted modules      & attn + MLP     & attn + MLP \\
    Learning rate        & $2\times10^{-5}$ & $1\times10^{-5}$ \\
    Epochs               & 3              & 2 \\
    Optimizer            & 8-bit AdamW    & 8-bit AdamW \\
    LR schedule          & cosine         & cosine \\
    Warmup               & $10\%$         & $10\%$ \\
    Weight decay         & 0.01           & 0.01 \\
    Gradient clipping    & 1.0            & 1.0 \\
    Effective batch size & 32             & 32 \\
    Max sequence length  & 1024           & 1024 \\
    Precision            & bfloat16       & bfloat16 \\
    Random seed          & 42             & 42 \\
    \bottomrule
  \end{tabular}
  \caption{Training hyperparameters for LoRA fine-tuning. The Gemma~3 recipe is held fixed
  across the 1B/4B/12B variants for the model-scaling study; TranslateGemma uses a lighter
  recipe to limit erosion of its pre-existing translation tuning. The effective batch size
  (32) is kept constant across model sizes by trading per-device batch against gradient
  accumulation.}
  \label{tab:hyperparams}
\end{table}
% TODO (collaborator): add NLLB-200-distilled-600M and Fairseq baseline training
% hyperparameters (optimizer, LR, batch size, steps/epochs); not yet on hand.

\section{Training Materials}\label{app:training_materials}
We have provided the translators with video recordings and guideline documents to familiarize with the task and to reference during the task. The guidleline document given for \Wikipedia{} domain English-Sinhala tranlsators is shown in Figures~\ref{fig:guidelines1}-\ref{fig:guidelines5}.

\newcommand{\guideImage}[1]{\begin{figure*}[t]
  \centering
  \includegraphics[width=\linewidth]{images/EnSi-Wikipedia_Parallel_Sentences_Cleaning-#1.pdf}
\caption{Guideline document given to the translators for the Wikipedia English-Sinhala task. - Page #1}
  \label{fig:guidelines#1}
\end{figure*}}

\guideImage{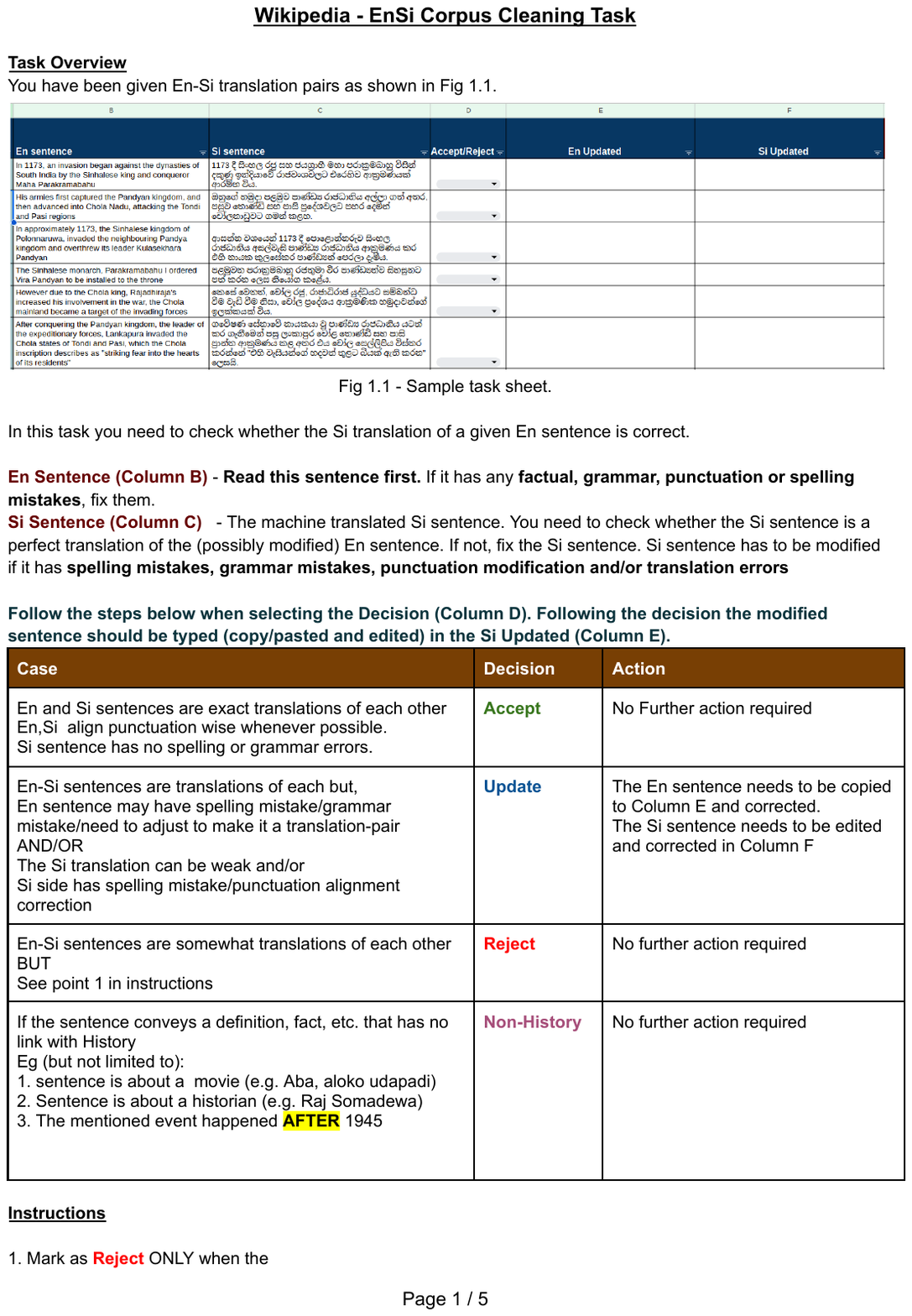}
\guideImage{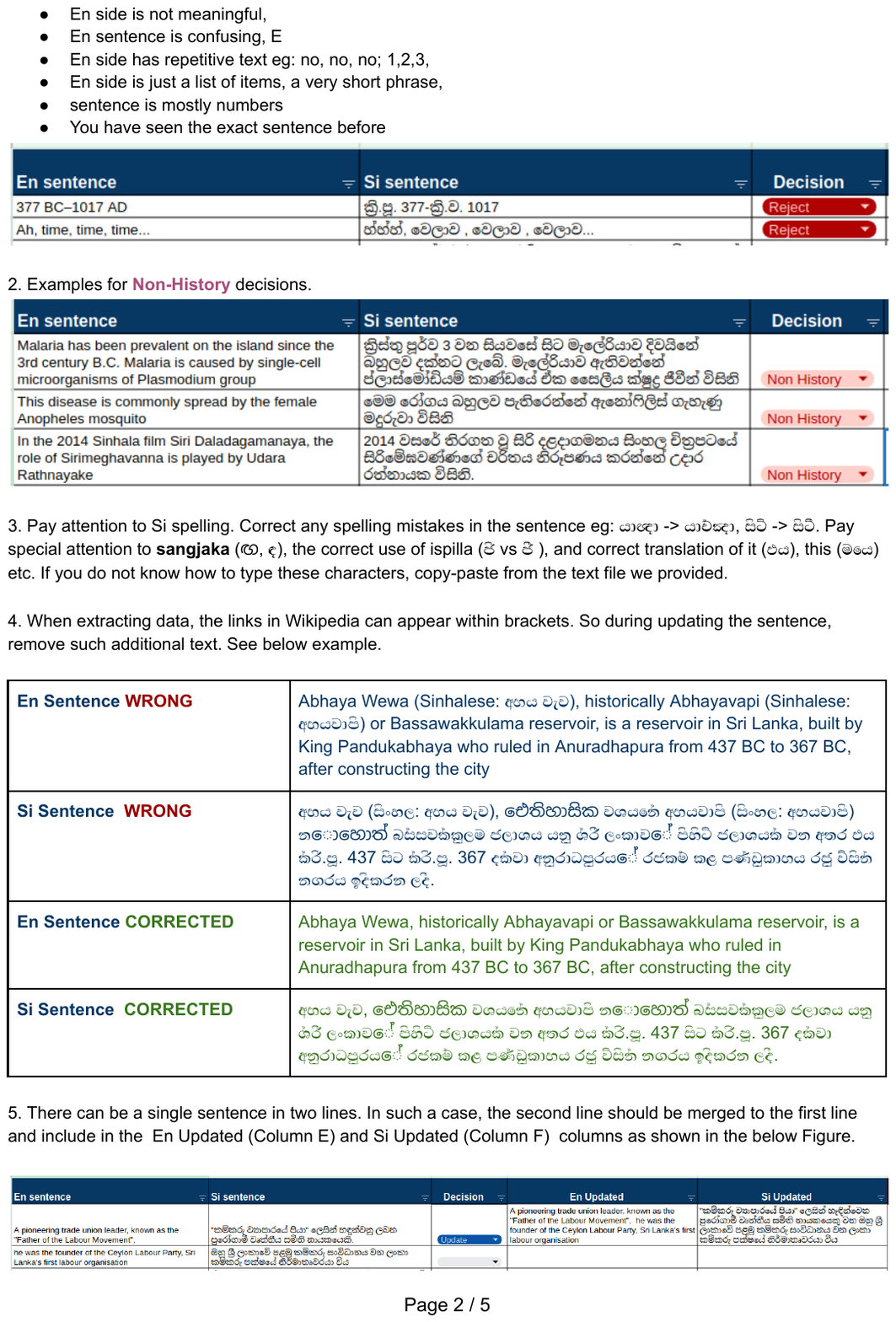}
\guideImage{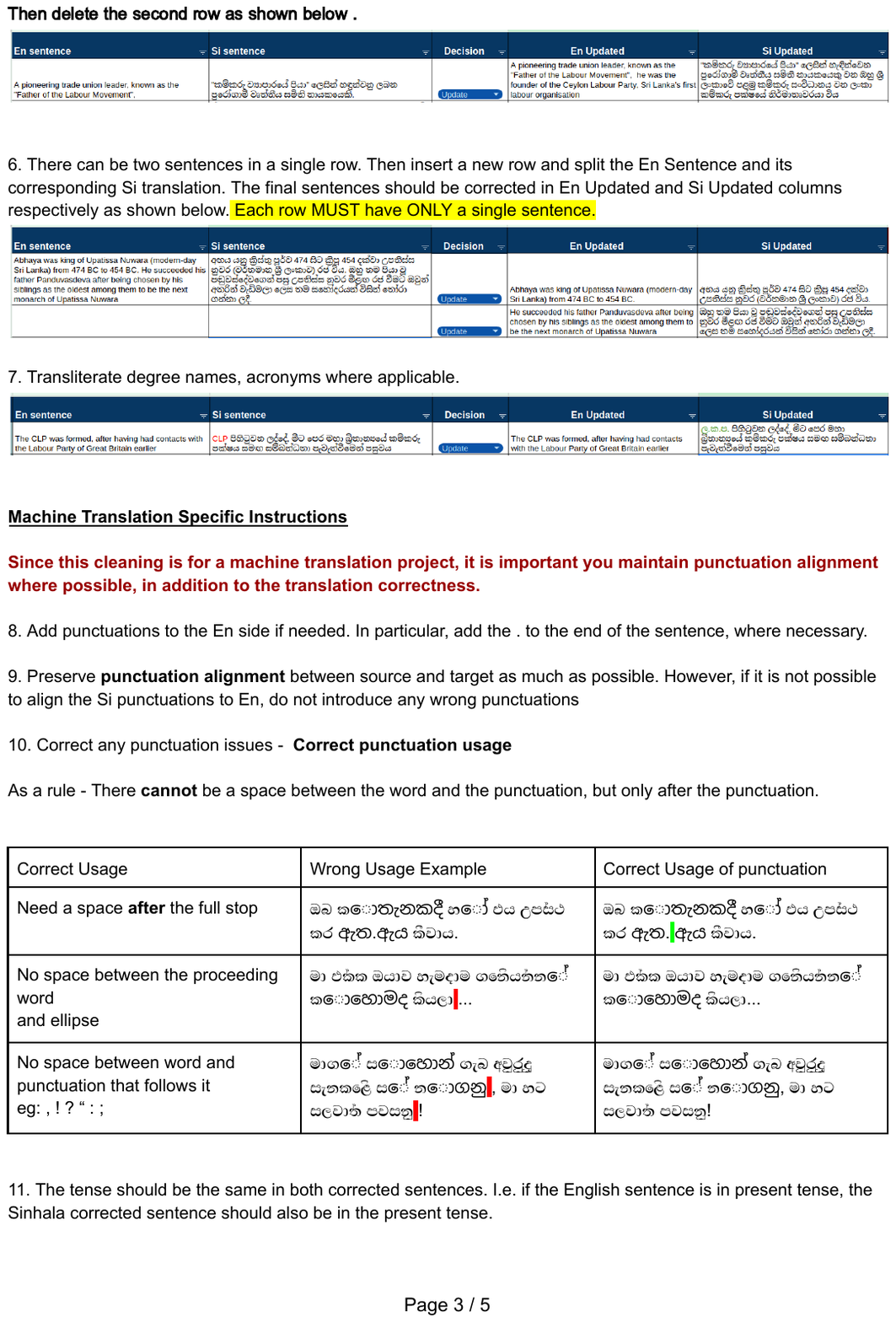}
\guideImage{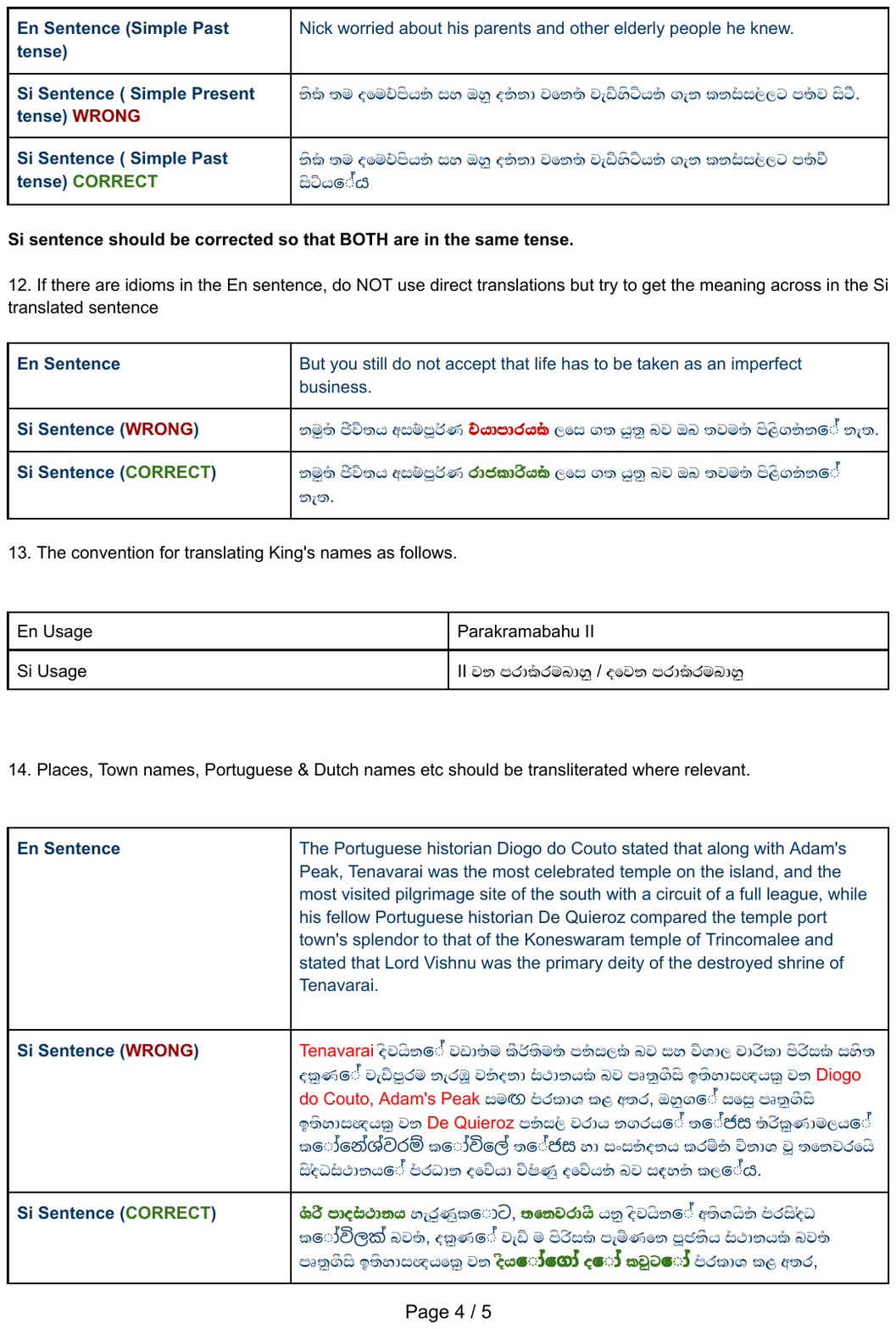}
\guideImage{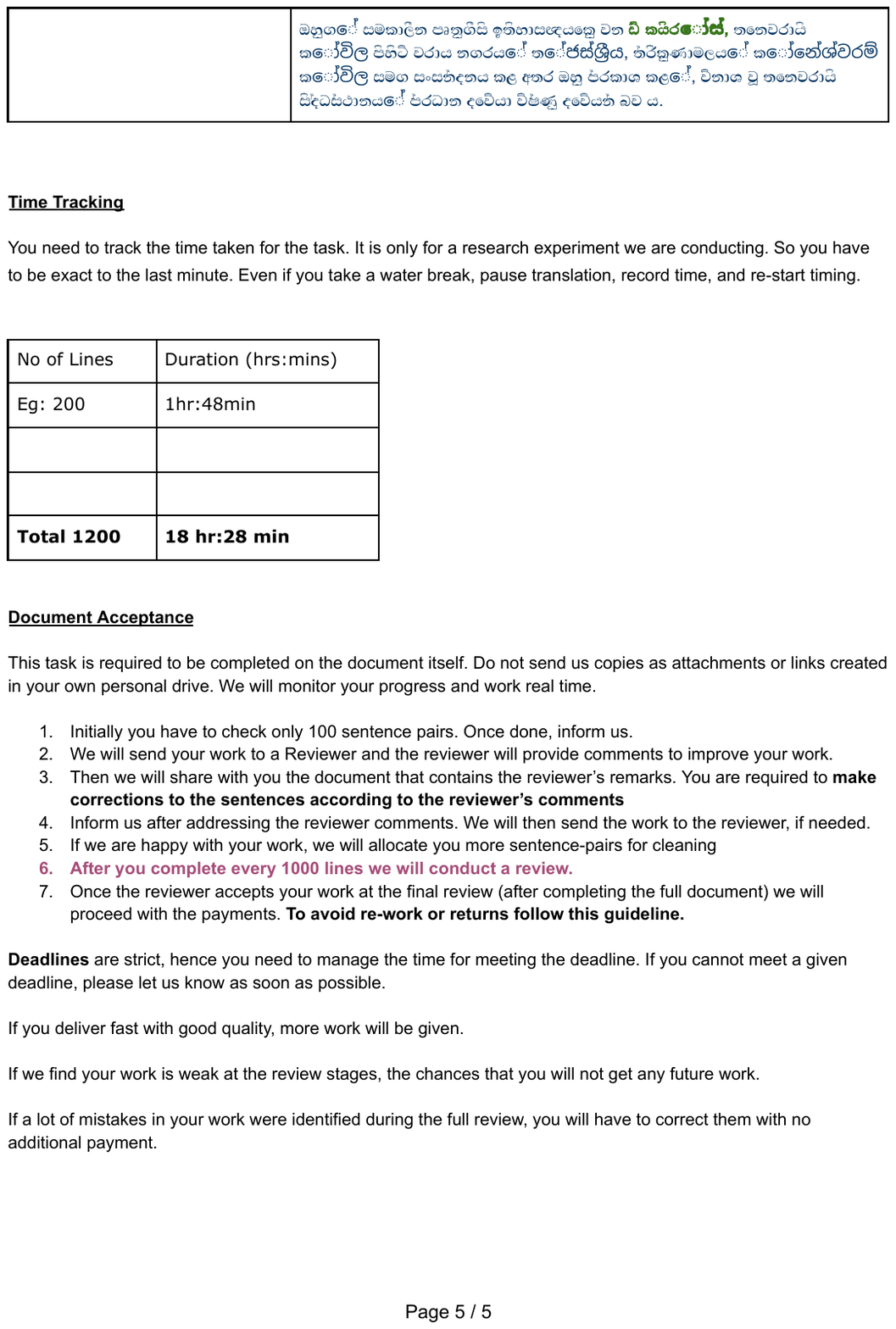}

%\section{Translator Details}\label{app:translator_details}

%In Table~\ref{tab:} section we provide an analysis based on translator's qualifications.

\end{document}